\documentclass{article}

\usepackage[preprint]{neurips_2025}

\usepackage[utf8]{inputenc} 
\usepackage[T1]{fontenc}    
\usepackage{hyperref}       
\usepackage{url}            
\usepackage{xurl} 
\usepackage{tabularx,booktabs}     
\usepackage{amsfonts}       
\usepackage{nicefrac}       
\usepackage{microtype}      
\usepackage{xcolor}         
\usepackage{tablefootnote}
\usepackage{diagbox}
\usepackage{makecell}
\usepackage{amsthm}
\usepackage{caption}
\usepackage{float}
\usepackage{graphicx}
\usepackage{multirow}
\usepackage{subcaption}
\usepackage{amssymb}
\usepackage{dsfont}
\usepackage{bbm}
\usepackage{amsmath}
\usepackage{algorithm}
\usepackage{algpseudocode}
\usepackage{physics}
\usepackage[normalem]{ulem}

\newcommand{\btheta}{\boldsymbol{\theta}}
\newcommand{\balpha}{\boldsymbol{\alpha}}
\usepackage{array}
\DeclareMathOperator{\argmin}{arg\,min}
\DeclareMathOperator{\argmax}{arg\,max}
\theoremstyle{plain}

\title{Merge to Mix: Mixing Datasets via Model Merging}

\author{%
  Zhixu Silvia Tao\thanks{Equal contribution. Authors listed in alphabetical order.} \thanks{Work was done while Zhixu was an intern at Fujitsu Research of America.}\\
  Princeton University
  \\
  \And
  Kasper Vinken\footnotemark[1] \\
  Fujitsu Research of America \\
  \AND
  Hao-Wei Yeh \\
  Fujitsu Limited\\
  \And
  Avi Cooper\\
  Fujitsu Research of America\\
  \And
  Xavier Boix \\
  Fujitsu Research of America
}

\begin{document}

\maketitle

\begin{abstract}

Mixing datasets for fine-tuning large models (LMs) has become critical for maximizing performance on downstream tasks. However, composing effective dataset mixtures typically relies on heuristics and trial-and-error, often requiring multiple fine-tuning runs to achieve the desired outcome. We propose a novel method, \textit{Merge to Mix}, that accelerates composing dataset mixtures through model merging. Model merging is a recent technique that combines the abilities of multiple individually fine-tuned LMs into a single LM by using a few simple arithmetic operations. Our key insight is that merging models individually fine-tuned on each dataset in a mixture can effectively serve as a surrogate for a model fine-tuned on the entire mixture. Merge to Mix leverages this insight to accelerate selecting dataset mixtures without requiring full fine-tuning on each candidate mixture. Our experiments demonstrate that Merge to Mix surpasses state-of-the-art methods in dataset selection for fine-tuning LMs.

\end{abstract}

\section{Introduction}
Fine-tuning large models (LMs) has emerged as an effective strategy for adapting LMs to diverse downstream tasks. The effectiveness of fine-tuning depends mostly on the quality of the training data distribution. Imbalanced or misaligned data distributions can lead to several issues such as degraded LM performance~\cite{iter2021complementarity, northcutt2021confident}, domain mismatches that hinder generalization~\cite{gururangan2020don}, etc. Thus, the composition of training data for fine-tuning has become critically important, shifting AI development from a model-centric to a data-centric approach~\cite{jarrahi2022principles, zha2025data}. 

Fine-tuning LMs requires data aligned with the target task. However, in practice, such data is often scarce or unavailable. A common and effective remedy is to mix multiple existing datasets during fine-tuning, leveraging the diversity of those datasets to obtain a training set more closely aligned with the target task. This approach introduces a challenging selection problem: identifying task-aligned dataset combinations among a large pool of candidate mixtures. As a result, selecting mixtures that align well with the target task often devolves into a costly trial-and-error process. Therefore, in this work we address the following problem:
\begin{quote} \emph{Given a target task, identify a mixture of available training datasets such that fine-tuning a pre-trained LM on this mixture yields the highest performance.} \end{quote}

We introduce \textit{Merge to Mix}, a novel method that leverages model merging for dataset mixing. Model merging \cite{matena2022merging, wortsman2022model, wortsman2022robust, ainsworth2022git, don2022cold, ilharco2022editing} is a recent technique that combines the capabilities of multiple LMs into a single LM without additional training cost. Given multiple LMs, each based on the same pre-trained LM but fine-tuned on a different dataset, model merging uses simple arithmetic operations such as averaging in the parameter space to integrate several fine-tuned models into a single model. 

Merge to Mix uses model merging to accelerate dataset mixture selection by eliminating the need to fine-tune for every candidate mixture. Our key insight is that the merged model can serve as an effective surrogate for a model fully fine-tuned on a mixture of datasets (referred to as \textit{mixture-fine-tuned} model). This enables efficiently exploring a large space of dataset mixtures without the need for fine-tuning to evaluate each mixture. Specifically, our contributions are as follows:
\begin{itemize}
    \item \textbf{Relation between merged and mixture-fine-tuned models:} We empirically show a strong positive correlation between the performance of a mixture-fine-tuned model and that of a merged model built from models individually fine-tuned on each dataset in the mixture. To the best of our knowledge, we are the first to introduce this correlation, opening a new direction for applying model merging.
    \item \textbf{Effective surrogate:}   Building upon this correlation, the merged model can be used as an effective surrogate to estimate the relative performance of mixture-fine-tuned models. 
    \item \textbf{Selection via surrogate:} Leveraging the surrogate, we propose \textit{Merge to Mix}, a novel method that uses the performance of merged models to guide and accelerate dataset mixture selection for fine-tuning LMs. 
    \item \textbf{Empirical validation:} Through a series of experiments on diverse target tasks and training datasets, we show that Merge to Mix consistently identifies high-performing dataset mixtures more efficiently and accurately than previous approaches, such as similarity-based selection.
\end{itemize}

\section{Related Work}\label{related_work}
We begin by reviewing prior work on data selection, the central focus of this paper. We then turn to model merging, a key technique employed in our proposed method, Merge to Mix.

\subsection{Data Selection}\label{sec:review_on_data_selection}

There is a growing interest in data selection due to its impact in the final accuracy of LMs. Most existing methods focus on sample-level selection, i.e., selecting individual data points for training based on their estimated quality or relevance \cite{xie2023data, ilyas2022datamodels, engstrom2024dsdm, iter2021complementarity, chen2023alpagasus, coleman2019selection}. However, with the proliferation of foundation models that are trained on massive corpora and support diverse downstream tasks, data selection has increasingly shifted to a higher dataset-level strategy, operating at the granularity of entire datasets \cite{xie2023doremi, fan2023doge, ye2024data, liu2024regmix}. Our work falls within this line of research, focusing on dataset-level selection by determining whether to include or exclude entire datasets.

Previous work has studied data selection across all stages of LM training \cite{albalak2024survey}, from pre-training \cite{chen2023skill, shen2023slimpajama, ye2024data, albalak2023efficient, xie2023data, xie2023doremi, feng2022automatic, engstrom2024dsdm, fan2023doge}, instruction tuning \cite{xia2024less, zhou2023lima, cao2023instruction, chen2023alpagasus, li2023quantity} to task-specific fine-tuning \cite{ivison2022data, lang2022training, maharana2023d2, phang2018sentence, iter2021complementarity, dong2024sketchy}. Our work falls in the latter category.
Unlike pre-training, which relies on large-scale and general-purpose data, fine-tuning requires task-specific data. Additionally, compared to pre-training, fine-tuning typically involves significantly shorter training times, as it already starts with a pre-trained, capable model. Though some techniques may be applicable to both pre-training and fine-tuning, these differences call for distinct selection strategies, and relatively little work has focused on dataset-level selection for fine-tuning.

Some related approaches include, but are not limited to, heuristic-based selection \cite{raffel2020exploring, laurenccon2022bigscience, longpre2023flan, phang2018sentence}, similarity-based selection \cite{aharoni2020unsupervised, xia2022moderate, ivison2022data, abbas2023semdedup}, proxy-model-based selection \cite{xie2023doremi, chen2023alpagasus, fan2023doge, coleman2019selection} and predictive-model-based selection \cite{aharoni2020unsupervised, iter2021complementarity, gunasekar2023textbooks, engstrom2024dsdm, liu2024regmix}. 
Merge to mix introduces a novel strategy that, to the best of our knowledge, does not fit existing categories. 
Similar to similarity-based and heuristic methods, Merge to Mix can operate without auxiliary models that may not always be available. However, Merge to Mix surpasses these approaches by using a more direct and principled signal: it directly uses task-performance feedback to guide dataset mixture selection.
Moreover, the primary operational advantage of Merge to Mix is the elimination of the fine-tuning step for evaluating a dataset mixture. This makes it complementary to approaches based on small proxy models or predictive  models, which both necessitate fine-tuning on candidate mixtures. By removing the fine-tuning requirement for mixture evaluation, Merge to Mix can be combined with these proxy-based and predictive-based techniques to further accelerate mixture selection, reducing both training and evaluation overhead.

\subsection{Model Merging}
As more fine-tuned LMs become publicly available, an interesting question is how to combine them into a single, general-purpose LM that contains all the abilities of the individually fine-tuned models. Model merging has emerged as a promising approach as it allows combining multiple LMs with just a few arithmetic operations \cite{matena2022merging, wortsman2022model, jin2022dataless, ainsworth2022git, don2022cold, yu2024language, ortiz2023task, stoica2023zipit, yadav2023ties, ilharco2022editing}. Recent work has investigated the underlying reasons behind the surprising effectiveness of model merging~\cite{tao2024task, ortiz2023task, ilharco2022editing, li2025task}. A variety of techniques have been developed to merge models, such as merging linear layers of fine-tuned models \cite{jin2022dataless}, merging with Task Arithmetic \cite{ilharco2022editing}, merging parameter-efficient modules \cite{zhang2023composing} and merging models trained from different initializations \cite{stoica2023zipit}. 

Model merging has been applied across a range of areas. For example, \cite{wortsman2022robust, wortsman2022model} used merging to improve model robustness and accuracy; \cite{izmailov2018averaging} averaged parameters of multiple models to enhance generalization; \cite{xiao2023lm} merged fine-tuned language models with pre-trained foundation models to mitigate catastrophic forgetting; \cite{liu2024towards} applied merging to unlearn harmful knowledge in LLMs; and \cite{abad2024strong} adaptively combined models to reduce reproduction of copyrighted content. While most prior work applies model merging during or after training, our approach introduces a novel use case: employing model merging at the data preparation stage for fine-tuning.

\section{Merge to Mix}\label{methodology}
In this section, we introduce Merge to Mix, a method for selecting dataset mixtures for fine-tuning an LM. We begin by formally defining the problem setting, then explain how model merging can be leveraged, and finally present the full algorithm.

\subsection{Notation and Problem Formulation}

Let $T$ be a target downstream task for which we want to maximize the accuracy of a pre-trained LM. We define $D_1, \dots, D_N$ as the available training datasets. Our goal is to select a mixture of these datasets to fine-tune the LM to maximize its performance on task $T$. We associate each dataset $D_i$ with a binary selection variable $\alpha_i \in \{0, 1\}$ to indicate whether to include the dataset for fine-tuning $(\alpha_i=1)$ or not $(\alpha_i=0)$, and $\balpha = (\alpha_1, \dots, \alpha_N)\in\{0, 1\}^N$ denotes the corresponding binary vector. We define the the selected dataset mixture as 
\begin{align}
    \mathcal{S}_{\balpha} = \{D_i \mid \alpha_i = 1, \ i = 1,\dots, N\}.
\end{align}
Let $\btheta_0\in \mathbb{R}^d$ be the parameters of the pre-trained LM.
We define $\btheta^*(\mathcal{S}_{\balpha})$ as the parameters of the mixture-fine-tuned LM on the selected $\mathcal{S}_{\balpha}$, starting from $\btheta_0$, i.e., 
\begin{align}
\btheta^*(\mathcal{S}_{\balpha}) \in \argmin_{\btheta\in \mathbb{R}^d}L(\btheta;\mathcal{S}_{\balpha}),
\end{align}
where $L(\cdot; \cdot)$ is the loss function, in which the first argument is the LM's parameters and the second is the dataset used for evaluating the loss. For simplicity, we omit in our notation through the paper that fine-tuning always starts from the same pre-trained LM with parameters $\btheta_0$.

Merge to Mix aims at optimizing ${\balpha}$ such that fine-tuning the LM on the selected mixture $\mathcal{S}_{\balpha}$ achieves the best possible performance on the target task $T$. This can be expressed as the following optimization problem:
\begin{align}\label{optimization_objective}
&\quad\argmin_{\balpha\in\{0, 1\}^N}L(\btheta^*(\mathcal{S}_{\balpha}); T)\nonumber\\
\textrm{s.t.}&\quad\btheta^*(\mathcal{S}_{\balpha})\in \argmin_{\btheta\in\mathbb{R}^d} L(\btheta; \mathcal{S}_{\balpha}).
\end{align}
Solving this optimization problem \eqref{optimization_objective} by exhaustively evaluating all $\balpha\in\{0, 1\}^N$ is computationally intractable in practice for two reasons. First, the size of the space $\{0, 1\}^N$ is $2^N$, which may be too large to evaluate for even a relatively small $N$. Second, since computing $\btheta^*(\mathcal{S}_{\balpha})$ requires one full fine-tuning run, identifying the optimal $\balpha$ by exploring $\{0, 1\}^N$ will incur a large number of fine-tuning runs. While search algorithms can reduce the number of evaluations needed, they still rely on fine-tuning, leaving the problem intractable at scale. Merge to Mix mitigates the cost of having to fine-tune many times by introducing a surrogate based on model merging.

Next, we introduce the surrogate to replace $\btheta^*(\mathcal{S}_{\balpha})$ in~\eqref{optimization_objective}, and later in Section~\ref{sec:algorithm}, we introduce the algorithm of Merge and Mix to optimize~\eqref{optimization_objective} by leveraging the surrogate.

\subsection{Model Merging as a Surrogate for \texorpdfstring{$\btheta^*(\mathcal{S}_{\balpha})$}{theta*(S_alpha)}}\label{sec:theory_of_model_merging}

To construct a surrogate for $\btheta^*(\mathcal{S}_{\balpha})$, we first define $\{\btheta^*_1, \dots, \btheta_N^*\}$ as the set of model parameters obtained by fine-tuning the pre-trained parameters $\btheta_0$ individually on each dataset in $\{D_1,\ldots,D_N\}$, i.e., $\btheta^*_i$ represents the model parameters fine-tuned on a single dataset $D_i$ such that $\btheta^*_i = \btheta^*(\{D_i\})$, for $i=1,\ldots,N$. We use $\tilde{\btheta}(\mathcal{S}_{\balpha})$ to denote the surrogate model, constructed by merging the models in $\{\btheta^*_1, \dots, \btheta^*_N\}$ corresponding to the datasets in $\mathcal{S}_{\balpha}$. To keep the process simple and avoid tuning additional hyperparameters, as required by other merging techniques such as \cite{ilharco2022editing, yu2024language}, we use averaging to merge the models. Thus, the surrogate $\tilde{\btheta}(\mathcal{S}_{\balpha})$ is defined in the following way: 
\begin{align}\label{formula_for_model_merging}
    \tilde\btheta(\mathcal{S}_{\balpha})=\frac{1}{|\mathcal{S}_{\balpha}|}\sum_{\{i: \alpha_i = 1\}}\btheta_i^*.
\end{align}
To use the surrogate, the pre-trained model $\btheta_0$ needs to be fine-tuned once per individual dataset in $\{D_1,\ldots,D_N\}$ to obtain $\{\btheta^*_1, \dots, \btheta^*_N\}$. These individually fine-tuned models can be pre-calculated and re-used for any evaluation of the surrogate $\tilde{\btheta}(\mathcal{S}_{\balpha})$. While the cost of $N$ fine-tuning runs is not negligible, it is relatively small: fine-tuning on one individual dataset tends to be less costly than fine-tuning on a mixture of datasets as the mixture usually involves more data. That is, with the surrogate, evaluating any mixture $\mathcal{S}_{\balpha}$ requires only an averaging of model parameters. Without the surrogate, each evaluation involves a full fine-tuning run. Thus, the surrogate enables exploring a much larger number of candidate mixtures $\balpha\in\{0,1\}^N$, leading to high final accuracy, as we show in the experiments section.

\emph{A priori}, it is unclear whether the surrogate $\tilde\btheta(\mathcal{S}_{\balpha})$ is effective to replace $\btheta^*(\mathcal{S}_{\balpha})$. Prior work has studied conditions under which model merging can effectively approximate fine-tuning~\cite{tao2024task, ortiz2023task, li2025task}. However, these conditions are typically restrictive, such as convex loss functions \cite{tao2024task} or single-headed and one-layer nonlinear Transformer models \cite{li2025task}. Without these conditions, merging may fail to approximate fine-tuning.

Though a close match between the merged model and the mixture-fine-tuned model would guarantee that $\tilde\btheta(\mathcal{S}_{\balpha})$ serves as an effective surrogate for $\btheta^*(\mathcal{S}_{\balpha})$, our approach requires substantially less. Specifically, given a target task $T$, we only require that the performance of the merged model $\tilde\btheta(\mathcal{S}_{\balpha})$ is positively correlated with that of the mixture-fine-tuned model $\btheta^*(\mathcal{S}_{\balpha})$, i.e., 
\begin{align}\label{eq:merged_loss_proportional_to_finetuned_loss}
L(\tilde\btheta(\mathcal{S}_{\balpha});T)\propto L(\btheta^*(\mathcal{S}_{\balpha}); T)\quad \forall \balpha, T,
\end{align}
Note that a direct consequence of this positive correlation is that 
\begin{align}\label{eq:solution_approximation}
       \balpha^* = \argmin L(\tilde\btheta(\mathcal{S}_{\balpha});T)\approx \argmin L(\btheta^*(\mathcal{S}_{\balpha});T),
\end{align} 
since the $\argmin$ is invariant to positive scalar scaling. In Section~\ref{sec:linear_correlation_exp}, we provide empirical evidence for Equation \eqref{eq:merged_loss_proportional_to_finetuned_loss}, demonstrating a high positive correlation between the performance of $\tilde\btheta(\mathcal{S}_{\balpha})$ and $\btheta^*(\mathcal{S}_{\balpha})$. To the best of our knowledge, our work is the first to explore this correlation-based view.

\begin{algorithm}[t]
\caption{Merge to Mix}
\label{alg:dataset_selection}
\begin{algorithmic}[1]
\State\textbf{Input:} Pre-trained model parameters $\btheta_0$, training datasets $\{D_1, \dots, D_N\}$, target dataset $T$, loss function $L$, number of epochs $E$, learning rate $\eta$, batch size $B$
\State\textbf{Output:} Optimal binary vector $\balpha^*$ and its corresponding selected set $\mathcal{S}_{\balpha^*}$ 
\For{each $n\in \{1, \dots, N\}$}
    \State $\btheta^*_i\gets$ Fine-tune $\btheta_0$ with $D_i, E, \eta, B$
\EndFor
\State Initialize $L_{\min} = \infty$ 
\For{each binary vector $\balpha\in\{0, 1\}^N$}
    \State $\tilde{\btheta}\gets \frac{1}{|\mathcal{S}_{\balpha}|}\sum_{i:\alpha_i=1}\btheta_i^*$
    \If{$L(\tilde\btheta;T)<L_{\min}$}
    \State $L_{\min} = L(\tilde\btheta; T)$, $\balpha^* = \balpha$ and $\mathcal{S}_{\balpha^*}= \{D_i: \alpha^*_i = 1\}$
    \EndIf
\EndFor
\State \Return $\balpha^*, \mathcal{S}_{\balpha^*}$
\end{algorithmic}
\end{algorithm}

\subsection{Algorithm}\label{sec:algorithm}

Leveraging the surrogate $\tilde\btheta(\mathcal{S}_{\balpha})$ introduced in the previous section, we now turn to the problem of optimizing \eqref{optimization_objective} over $\balpha \in \{0,1\}^N$. By substituting the surrogate, we can reformulate the original optimization problem \eqref{optimization_objective} as the following: 
\begin{align}\label{optimization_objective_for_model_merging}
    &\quad\argmin_{\balpha\in\{0,1\}^N}L(\tilde\btheta(\mathcal{S}_{\balpha}); T)\nonumber\\
    \textrm{s.t.}&\quad\tilde\btheta(\mathcal{S}_{\balpha}) = \frac{1}{|\mathcal{S}_{\balpha}|}\sum_{\{i:\alpha_i = 1\}}\btheta^*_i.
\end{align}
As mentioned earlier, a naive approach would require up to $2^N$ full fine-tuning runs to evaluate all $\balpha\in\{0, 1\}^N$. However, this new formulation with the surrogate enables efficient evaluation of candidate $\balpha$ and tractable search for high-performing mixtures, all without additional training cost. Now, we present our algorithm Merge to Mix for solving the surrogate optimization problem (\ref{optimization_objective_for_model_merging}). Merge to Mix (Algorithm \ref{alg:dataset_selection}) consists of two key steps: (1) fine-tune $N$ individual models $\{\btheta_1^*, \dots, \btheta_N^*\}$ on $\{D_1, \dots, D_N\}$ , followed by (2) merging and evaluating the merged models on the target task $T$ to identify the optimal mixture of datasets. We detail each step below.

\textbf{Step 1: Fine-tune $N$ models.} For each available training dataset $D_i$, we fine-tune the model on $D_i$ starting from the same pre-trained model $\btheta_0$, using the same learning rate $\eta$, batch size $B$ and number of epochs $E$. The choice of using the same hyperparameter setup for all fine-tuning follows the previous work \cite{tao2024task} which showed that uniform fine-tuning improves model merging performance.

\textbf{Step 2: Merge and evaluate the merged models.} For each binary vector $
\balpha\in\{0,1\}^N$, we merge the fine-tuned models corresponding to the selected datasets (i.e., those with $\alpha_i=1$), using simple averaging as described in Equation~\eqref{formula_for_model_merging}. The merged model is then evaluated on the target task $T$. We repeat this process across all binary vectors and select $\balpha^*$ that yields the best-performance merged model on $T$. Thanks to the efficiency of model merging, we were able to exhaustively evaluate all candidate mixtures for the sizes of $N$ considered in the experiment section. As a result of exhaustively exploring all candidate mixtures, $\balpha^*$ is the optimal solution of problem \eqref{optimization_objective_for_model_merging}, and thus, conditional on \eqref{eq:merged_loss_proportional_to_finetuned_loss}, approximates the optimal solution of the original problem (\ref{optimization_objective}) (See Equation (\ref{eq:solution_approximation})).
The selected subset $\mathcal{S}_{\balpha^*} := \{D_i \mid \alpha_i^* = 1,\ i = 1,\dots, N\}$ constitutes the final choice of dataset mixture for fine-tuning on the target task $T$. 

Merge to Mix can be extended to more general settings. First, when the search space is substantially larger due to high $N$, Merge to Mix can be combined with search algorithms or predictive models to reduce the number of evaluations by prioritizing promising candidates. Second, weighted dataset mixing, which allows $\balpha\in[0,1]^N$ instead of restricting to $\{0, 1\}^N$ and has been considered in the literature (e.g.,\cite{xie2023doremi}), can also be implemented by leveraging alternative merging techniques such as weighted averaging. Although we do not empirically investigate these settings in our study, Merge to Mix can, in principle, be easily extended to support them, as discussed in Section~\ref{conclusion}.


\section{Experiments}\label{experiments}
In this section, we begin by validating the central assumption of Merge to Mix from Section \ref{sec:theory_of_model_merging}: that the merged model performance correlates strongly with the performance of the models fine-tuned on the corresponding dataset mixture. Then, we demonstrate that this correlation enables Merge to Mix to achieve superior dataset selection performance. To show the generality of this approach, we present experiments covering both computer vision and language tasks.

\subsection{Setup}\label{experimental_setup}
\textbf{Computer vision tasks and model.} First, we evaluated our approach on eight computer vision image classification datasets: Cars \cite{krause20133d}, DTD \cite{cimpoi2014describing}, EuroSAT \cite{helber2019eurosat}, GTSRB \cite{stallkamp2011german}, MNIST \cite{lecun1998mnist}, RESISC45 \cite{cheng2017remote}, SUN397 \cite{xiao2016sun}, and SVHN \cite{netzer2011reading}. While each of these datasets has its own training set, in real-world scenarios new computer vision problems do not typically come with dedicated labeled data. To simulate this, we performed leave-one-task-out evaluation: in each of eight experiments, one dataset was designated as the held-out target task, whereas the remaining seven were treated as available training data. The goal was to select the best mixture of datasets from the seven training datasets. We then evaluated the selected mixture on the held-out dataset. We used CLIP-ViT-B-32 \cite{radford2021learning} as our pre-trained model. We fine-tuned it on each training dataset individually for 10 epochs with batch size 128. For more details on the fine-tuning process, please refer to Appendix \ref{appendix:details_on_fine_tuning}.

\textbf{Language tasks and model.} Second, to further assess the generality of our approach beyond vision, we extended our experiments to the language domain. Specifically, we fine-tuned a large language model to improve non-English general language understanding skills, using Japanese as a representative case. We used six training datasets: mCoT-MATH-ja \cite{lai2024mcot}, OASST2-33k-ja, OASST1-21k-en, Aya-dataset-Japanese \cite{singh2024aya}, Japanese/English translations and a subset of English samples and machine-translated-to-Japanese samples from SlimOrca \cite{SlimOrca}. We used a battery of benchmarks with different metrics to evaluate the following eight Japanese skills, in addition to English understanding: translation (from and to English), information extraction, mathematical reasoning, entity extraction, question answering, semantic analysis, syntactic analysis, and commonsense morality understanding. For more details on datasets and metrics, please refer to Appendix \ref{appendix:ds_and_eval_for_language_tasks}. We used Llama-3-8B-Instruct \cite{llama3modelcard} as our pre-trained model, which was pretrained primarily on English data. We fine-tuned it on each training dataset individually for 1 epoch with learning rate 0.01. The fine-tuning is implemented with IA3 \cite{liu2022few}, a parameter-efficient fine-tuning technique. For more details on the fine-tuning process, please refer to Appendix \ref{appendix:details_on_fine_tuning}.


\begin{figure}[t]
    \begin{subfigure}{.5\textwidth}
    \centering
    \includegraphics[width =0.9\textwidth]{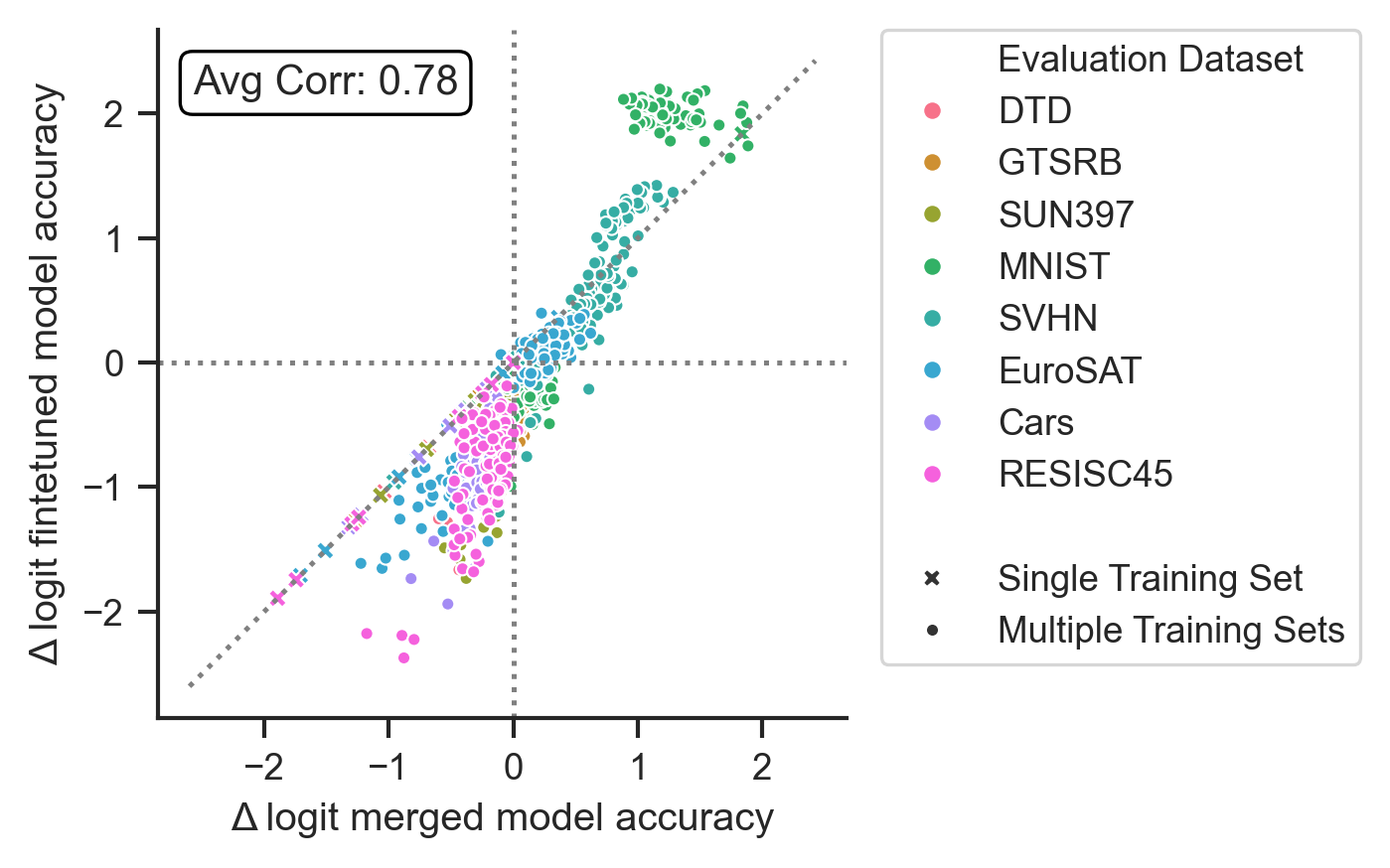}
    \caption{Image Classification: Merged Model}
    \label{fig:vision_correlation}
    \end{subfigure}
    \begin{subfigure}{.5\textwidth}
    \centering
    \includegraphics[width = 0.9\textwidth]{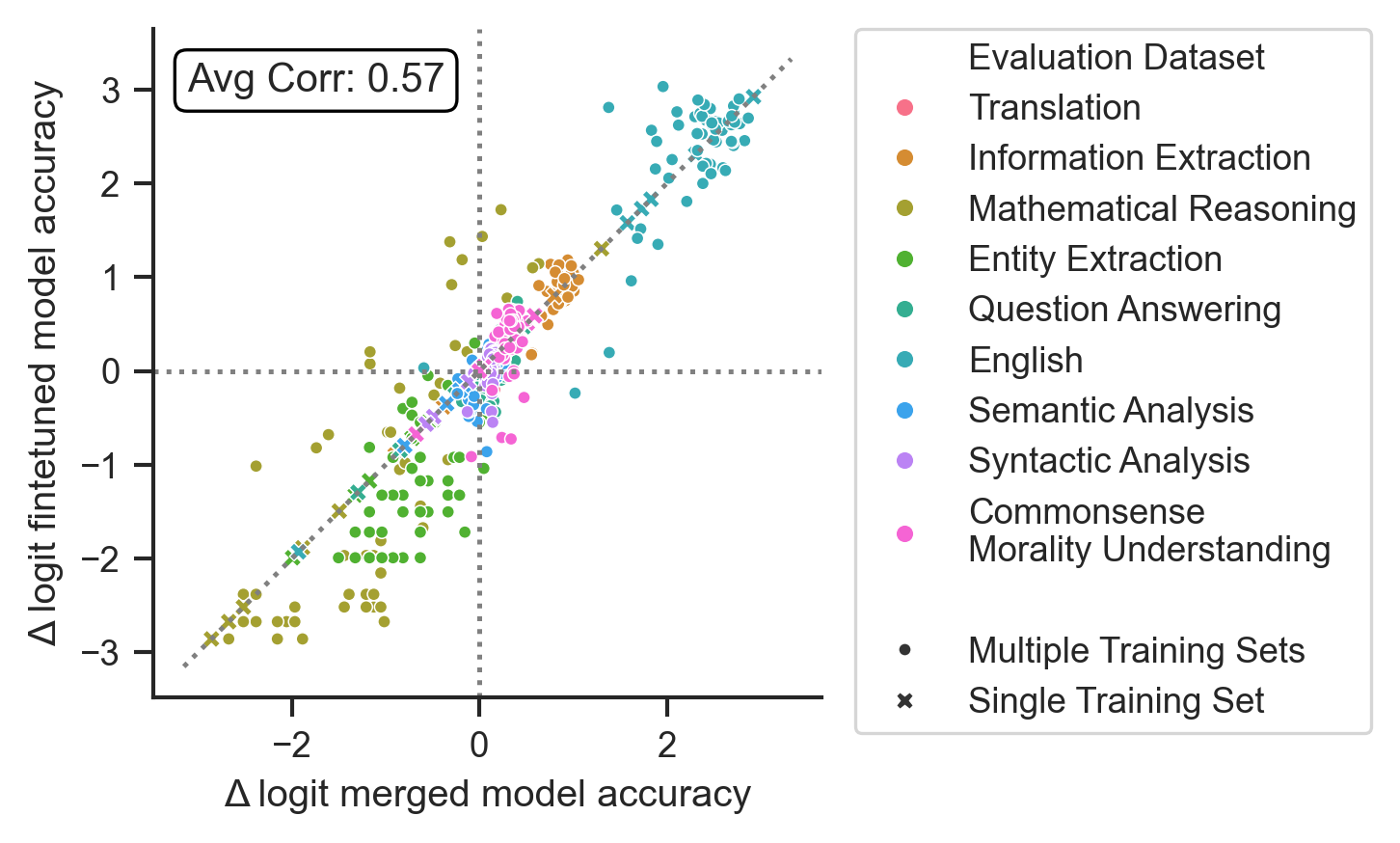}
    \caption{Language: Merged Model}
\label{fig:language_correlation}
    \end{subfigure}
    \begin{subfigure}{.5\textwidth}
    \centering
    \includegraphics[width =0.9\textwidth]{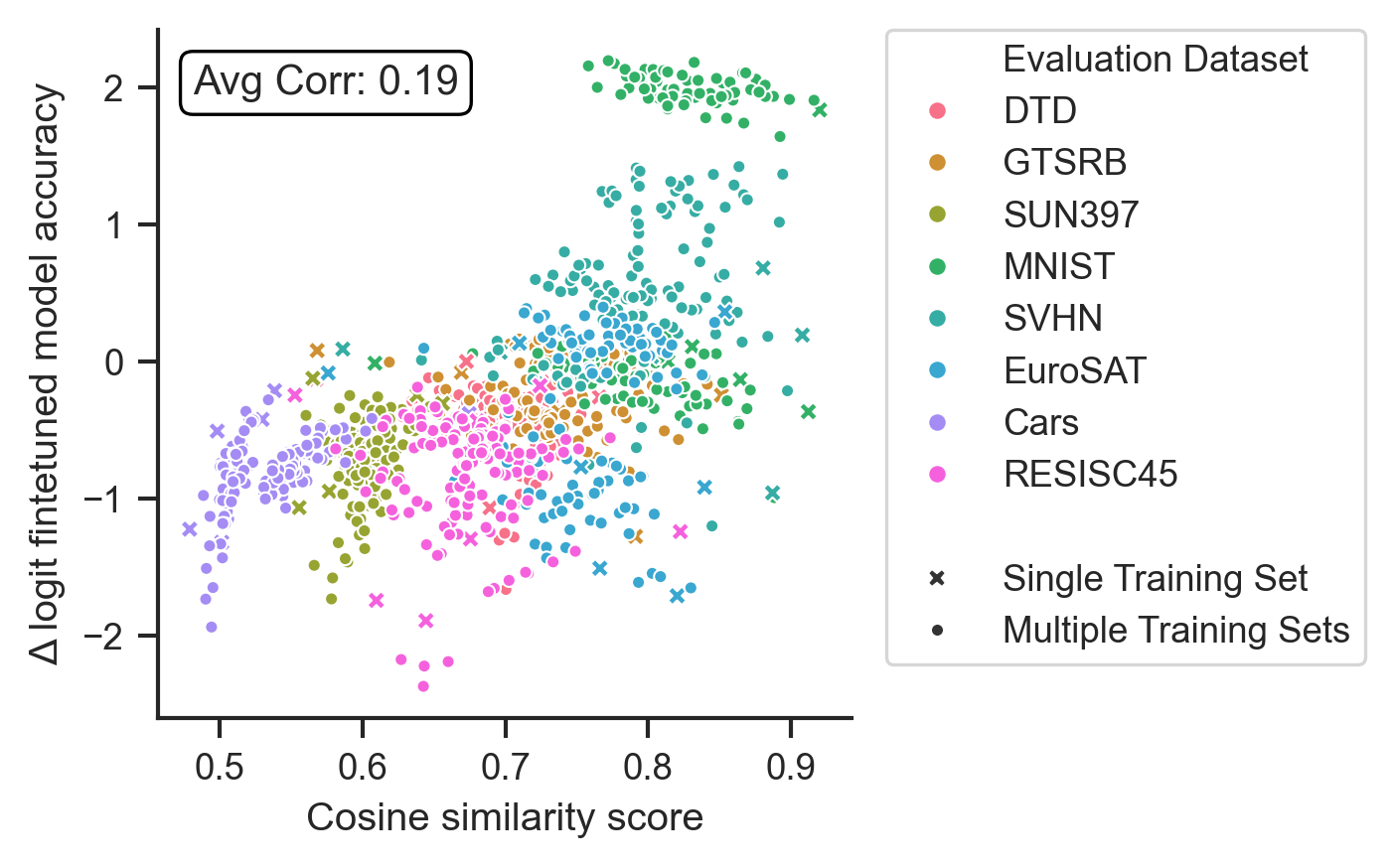}
    \caption{Image Classification: Cosine Similarity}
    \label{fig:cos_correlation_vision}
    \end{subfigure}
    \hfill
    \begin{subfigure}{.5\textwidth}
    \centering
    \includegraphics[width =0.9\textwidth]{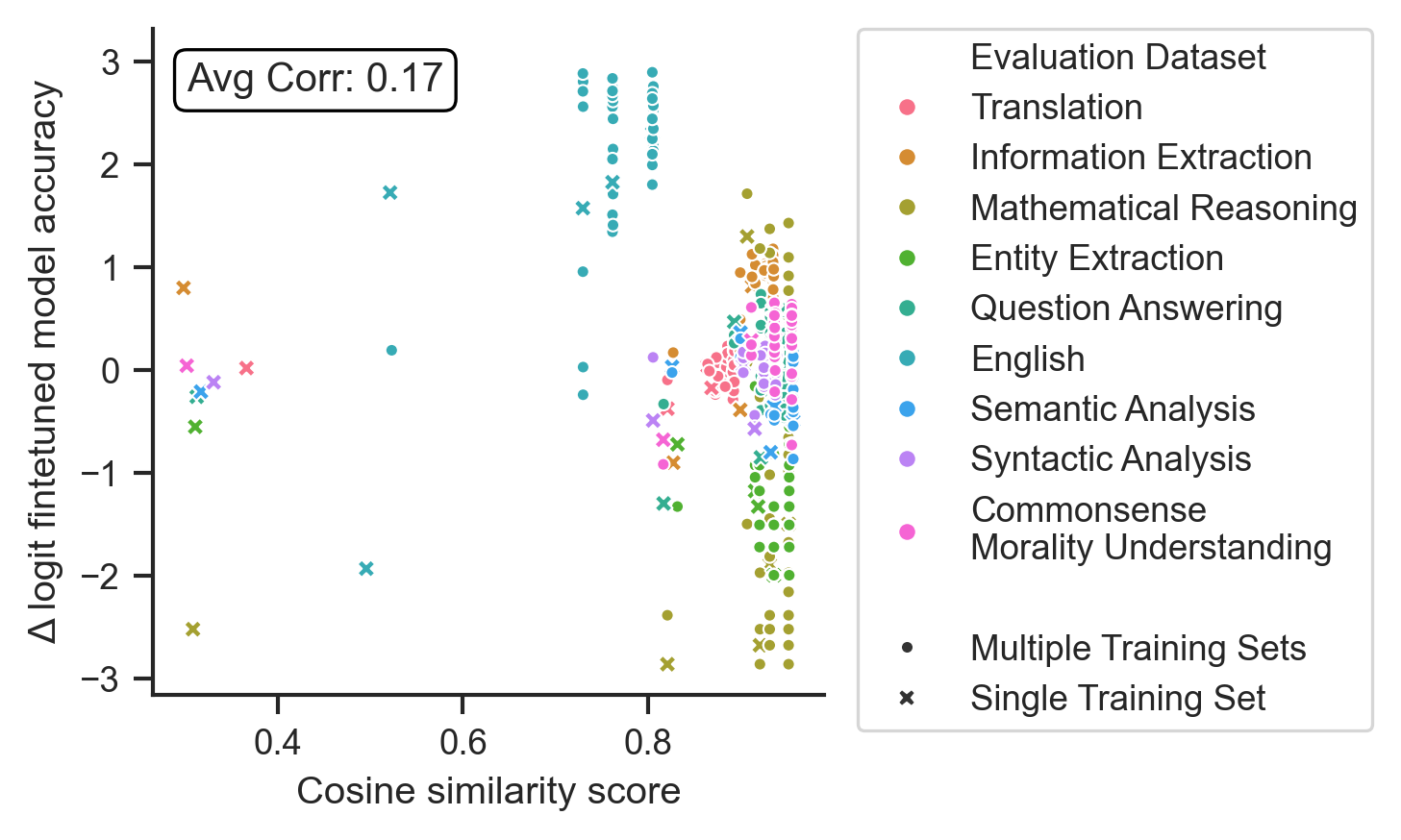}
    \caption{Language: Cosine Similarity}
    \label{fig:cos_correlation_language}
    \end{subfigure}
    \begin{subfigure}{.5\textwidth}
    \centering
    \includegraphics[width = 0.9\textwidth]{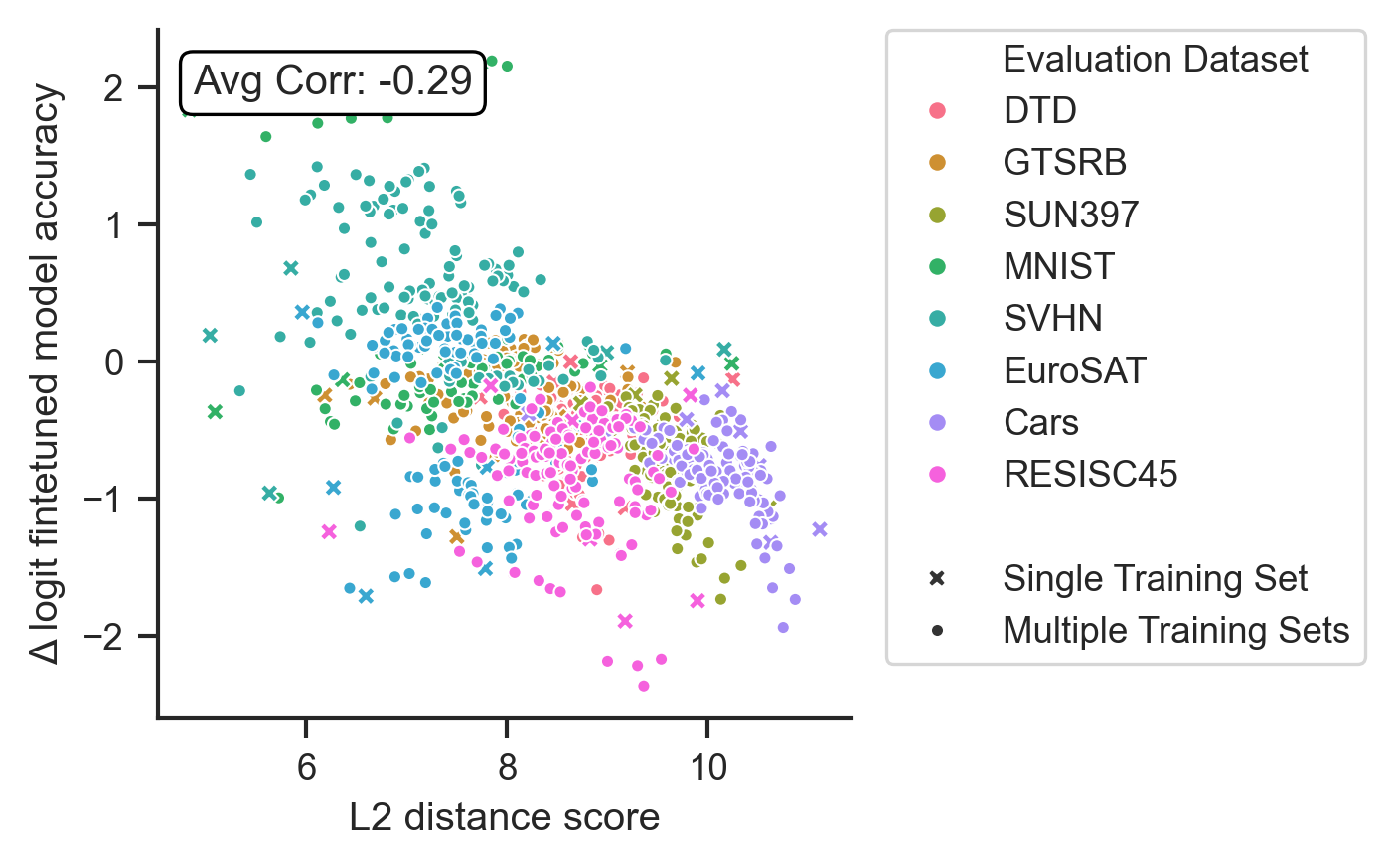}
    \caption{Image Classification: $L_2$ Score}
    \label{fig:L2_correlation_vision}
    \end{subfigure}
    \hfil
    \begin{subfigure}{.5\textwidth}
    \centering
    \includegraphics[width = 0.9\textwidth]{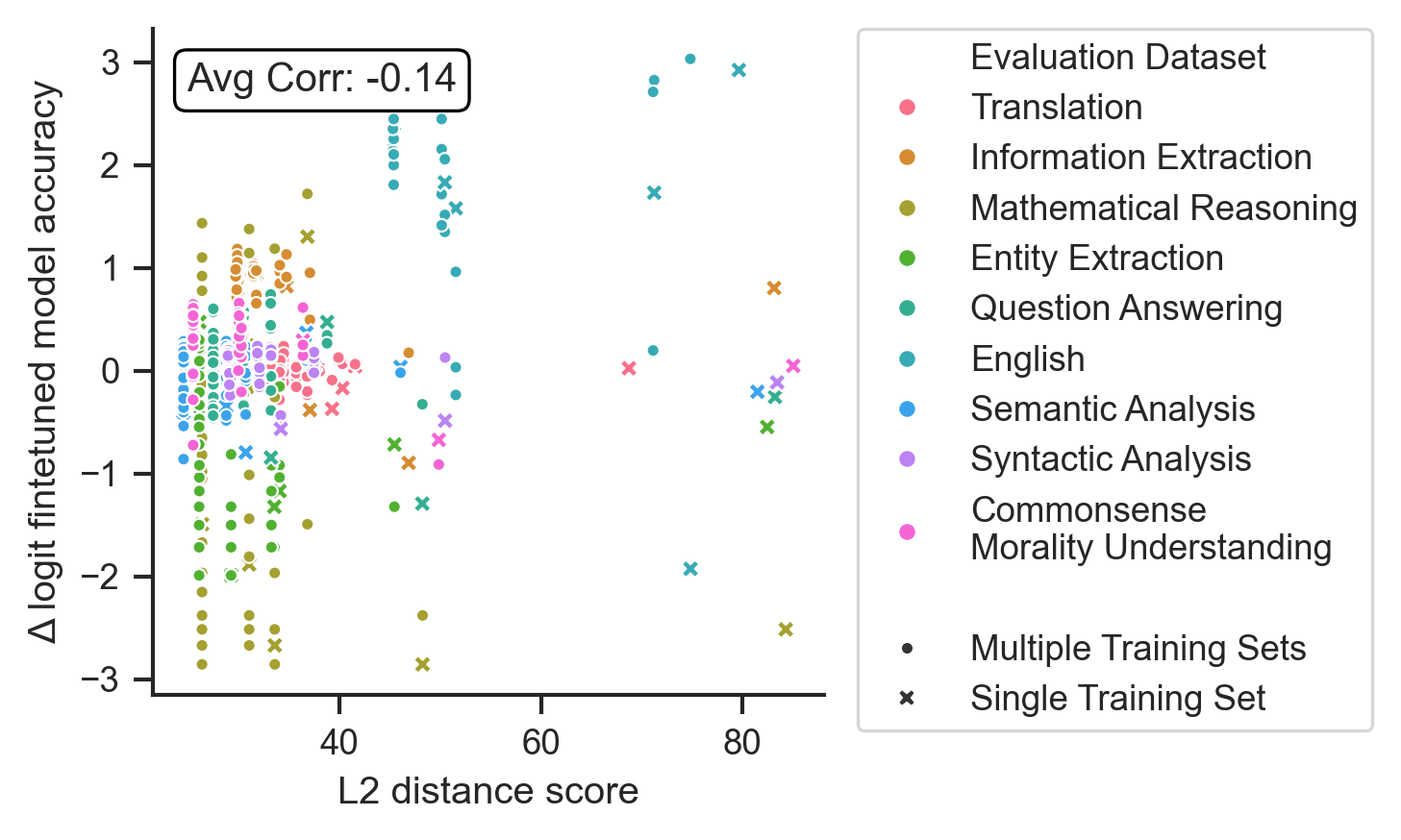}
    \caption{Language: $L_2$ Score}
    \label{fig:L2_correlation_language}
    \end{subfigure}    
\caption{Correlation plots between the performance of mixture-fine-tuned models and different metrics on target tasks. The reported correlation is the average of per-dataset-correlation.}
\label{fig:correlation}
\end{figure}

\textbf{Baselines.}  We considered three baselines: fine-tuning on all available datasets, similarity-based selection (using the embeddings of the pre-trained model) and random selection. Fine-tuning on all training datasets reflects a common heuristic: using more data should improve generalization. For instance, \cite{radford2019language} showed that language models trained on massive datasets like WebText exhibit strong zero-shot performance across diverse tasks. 

Similarity-based selection serves as a key baseline because, like Merge to Mix, it does not require any auxiliary models, making it applicable in scenarios where such models are unavailable or unreliable. For similarity-based selection, we adopted a total of six variants based on cosine similarity and $L_2$ distance, allowing the selection method to benefit from a wide range of similarity formulations. Detailed formulations and implementation can be found in Appendix \ref{appendix:similarity_based_selection}. 

Random selection represents the expected test accuracy if a datasets mixture were chosen uniformly at random from all possible mixtures of available datasets. It provides a non-heuristic baseline, as it does not incorporate any prior assumptions or selection criteria. Formally, let $\operatorname{Acc}(\btheta;T)$ denote the accuracy of a model with parameters $\btheta$ on task $T$. We report the performance of random selection as the average test accuracy across models fine-tuned on all $2^N-1$ possible mixtures of $N$ datasets excluding the empty set:
$\frac{1}{2^N-1}\sum_{\balpha}\operatorname{Acc}(\btheta^*(\mathcal{S_{\balpha}});T)$.

\subsection{Correlation between Accuracy of Mixture-Fine-Tuned Models and Merged Models}\label{sec:linear_correlation_exp}

To provide empirical evidence for the correlation hypothesized in Section~\ref{sec:theory_of_model_merging}, we analyze the correlation between the test accuracy of merged models $\tilde\btheta$ on the target task and that of mixture-fine-tuned models $\btheta^*$. For comparison, we also analyze the correlation between similarity-based metrics and the mixture-fine-tuned model accuracy. All correlational results are reported in Figure \ref{fig:correlation}. For all the plots in Figure
\ref{fig:correlation}, each color represents a different target task. For each task, there are $N$ candidate training datasets from which selections can be made, yielding $2^N-1$ points per color. Each point corresponds to a non-empty set of selected datasets $\mathcal{S}_{\balpha}$, specified by a binary vector $\balpha$. The empty set (i.e., selecting no dataset) is excluded. Moreover, the $y$-axis always shows the improvement of the mixture-fine-tuned model over the pre-trained model, measured as a change in logit accuracy. That is, using $\operatorname{Acc}(\btheta;T)$ to denote the accuracy of a model with parameters $\btheta$ on task $T$, we have $y = \operatorname{logit}(\operatorname{Acc}(\btheta^*(\mathcal{S}_{\balpha});T)) - \operatorname{logit}(\operatorname{Acc}(\btheta_0;T))$ for some $\balpha$ and $T$, where $\operatorname{logit}(p) = \log(\frac{p}{1-p})$ transforms the accuracy into log-odds.

\textbf{There was a strong linear correlation between the accuracy of the merged and mixture-fine-tuned models.} Figures \ref{fig:vision_correlation} and \ref{fig:language_correlation} show how closely the accuracy of merged models align with mixture-fine-tuned models, corresponding to image classification tasks and language tasks, respectively. The x-axis shows how much the merged model improves over the pre-trained model on the target task, measured as a change in logit accuracy. That is,
    $x = \operatorname{logit}(\operatorname{Acc}(\tilde\btheta(\mathcal{S}_{\balpha}); T)) - \operatorname{logit}(\operatorname{Acc}(\btheta_0;T))$. In both figures, a cross mark denotes mixtures with single dataset, where merged and mixture-fine-tuned models are identical, so it lies on the line $x = y$. A dot indicates mixtures with multiple datasets. 
    
To quantify this relation, we report the average correlation across all evaluation tasks in Figures \ref{fig:vision_correlation} and \ref{fig:language_correlation}. Specifically, for each task $T$, we computed the Pearson correlation coefficient for the set of pairs $\{(\operatorname{Acc}(\tilde\btheta(S_{\balpha});T), \operatorname{Acc}(\btheta^*(\mathcal{S}_{\balpha});T))\}_{\balpha\in \{0, 1\}^N}$, i.e., between the merged model and mixture-fine-tuned model test accuracy. We excluded the cases where $\operatorname{Acc}(\tilde\btheta(S_{\balpha});T) = \operatorname{Acc}(\btheta^*(\mathcal{S}_{\balpha});T)$, which occurs when fine-tuning is performed on a single dataset, resulting in  $\btheta^*(S_{\balpha}) = \tilde\btheta(S_{\balpha})$, to avoid inflating the correlation values. We computed the correlations separately for each target task and then averaged across all tasks to minimize any potential inflation from between-task variation. We observed a clear and strong positive correlation between the accuracy of mixture-fine-tuned models and merged models. Specifically, the average Pearson correlation coefficient was 
0.78 for the image classification tasks (Figure~\ref{fig:vision_correlation}), and 
0.57 for the language tasks (Figure~\ref{fig:language_correlation}). This provides empirical evidence for Equation (\ref{eq:merged_loss_proportional_to_finetuned_loss}), indicating that the performance of the merged model can effectively approximate the performance of the mixture-fine-tuned model. For the correlation plots and Pearson correlation coefficients separately per evaluation task, please refer to Appendix \ref{appendix:correlation_plots}.

\textbf{No strong correlation was observed between the performance of mixture-fine-tuned models and similarity-based metrics.} Figures \ref{fig:cos_correlation_vision} to \ref{fig:L2_correlation_language} show the correlation between similarity scores, computed for pairs of target tasks and dataset mixtures, and the accuracy of corresponding mixture-fine-tuned models. In this section, we present results for two of the six similarity-based metrics; the remaining four are provided in Appendix~\ref{appendix:correlation_plots}. Cosine similarity scores (Figures~\ref{fig:cos_correlation_vision} and~\ref{fig:cos_correlation_language}) were computed using Equation~\eqref{avg_of_max_cos}, while 
$L_2$ distance scores (Figures~\ref{fig:L2_correlation_vision} and~\ref{fig:L2_correlation_language}) were computed using Equation~\eqref{avg_of_min_L2}. 

Intuitively, a higher cosine similarity score or a lower $L_2$ distance score should suggest that the target task is more similar to a given dataset mixture, and therefore, the model fine-tuned on that mixture is expected to perform well on the target task. However, as shown in Figures \ref{fig:cos_correlation_vision} and \ref{fig:cos_correlation_language}, the average Pearson correlation for cosine similarity was only 0.19 for the image classification and 0.17 for the language tasks. Figures \ref{fig:L2_correlation_vision} and \ref{fig:L2_correlation_language} show that, for the $L_2$ distance, these values were -0.29 and -0.14, respectively. These correlations are significantly weaker than the 0.78 and 0.57 observed from model merging for image classification and language tasks respectively. These results challenge the common assumption that similarity-based methods can identify the most effective datasets mixture for training, which was also observed and reported in \cite{engstrom2024dsdm}.

\subsection{Merge to Mix Yields High-Performance Datasets Mixtures}

In this section, we evaluate the performance of Merge to Mix and compare it with the three baseline methods discussed in Section \ref{experimental_setup}: fine-tuning on all datasets, similarity-based selection and random selection. In addition to the baseline methods, we report an \textit{oracle} accuracy. Oracle reflects the best possible outcome under exhaustive fine-tuning, providing a meaningful performance ceiling. Oracle values were computed by selecting the dataset mixture that yields the highest validation accuracy on the target task. Specifically, we report
$\operatorname{Acc}(\btheta^*(\mathcal{S}_{\balpha^{\operatorname{oracle}}}); T)$ such that $\balpha^{\operatorname{oracle}} = \argmax_{\balpha}\operatorname{Acc}(\btheta^*(\mathcal{S}_{\balpha}); T^{\operatorname{val}})$.

Table \ref{tab:vision_accuracy} and \ref{tab:language_accuracy} correspond to image classification and language tasks, respectively. All reported values are from independent test splits. The corresponding validation splits were used to select a candidate according to the each selection method's criterion: maximum accuracy for Merge to Mix and for oracle, and maximum similarity or minimum distance for Similarity Selection. For tasks that did not contain a validation split, we constructed train-validation splits from the original training data. For all methods Tables \ref{tab:vision_accuracy} and \ref{tab:language_accuracy}, we included all mixtures with one or more datasets as potential candidates.

\begin{table}[t]
    \centering
    \resizebox{1.0\textwidth}{!}{%
    \begin{tabular}{cccccc|c}
        \toprule
        \diagbox{\textbf{Target Task}}{\textbf{Selection Method}}
 & \makecell{\textbf{Merge to Mix}\\(Merged Model)} &\makecell{\textbf{Merge to Mix}\\(Fine-tuned Model)} & \textbf{All Datasets}& \makecell{\textbf{Similarity}\\\textbf{Selection}}&\makecell{\textbf{Random}\\\textbf{Selection}} &\textbf{Oracle}\\
        \midrule
        Cars  &  \textbf{0.578}  & 0.428  &0.351  &0.456 &0.395 &0.544 \\
        DTD   &  \textbf{0.423} & 0.4 & 0.334  & 0.332 &0.334 &  0.393\\
        EuroSAT  & \textbf{0.606} & 0.511 & 0.469 & 0.429 &0.371& 0.536\\
        GTSRB &\textbf{ 0.393} & 0.334&  0.326& 0.309&0.284& 0.361 \\
       MNIST  & 0.855 & 0.88 &  \textbf{0.885}& 0.863&0.66 & 0.893\\
        RESISC45  &\textbf{0.599} & 0.445 &  0.415& 0.445&0.399 & 0.56\\
        SUN397  &\textbf{0.619} & 0.546 &  0.434& 0.469 &0.456 & 0.603\\
        SVHN & 0.625 &\textbf{0.645} &  0.558& 0.561&0.437 & 0.657\\\midrule
        Average & \textbf{0.587} & 0.524& 0.482 &0.483 &0.417& 0.568 \\
        \bottomrule\\
    \end{tabular}}
    \caption{Performance of Different Dataset Selection Methods on Image Classification Tasks}
    \label{tab:vision_accuracy}
\end{table}
\begin{table}[t]
    \centering
    \resizebox{1.0\textwidth}{!}{%
    \begin{tabular}{cccccc|c}
        \toprule
\diagbox{\textbf{Target Task}}{\textbf{Selection Method}}
  & \makecell{\textbf{Merge to Mix}\\(Merged Model)} &\makecell{\textbf{Merge to Mix}\\(Fine-tuned Model)} & \textbf{All Datasets}&\makecell{\textbf{Similarity}\\\textbf{Selection}}  &\makecell{\textbf{Random}\\\textbf{Selection}} &\textbf{Oracle}\\
        \midrule
        Translation  & \textbf{0.211} & 0.176  & 0.197 & 0.184&0.179  &0.216 \\
        Information Extraction  & 0.851 & \textbf{0.856}& 0.839 & 0.847&0.822  &0.867\\
        Mathematical Reasoning  & \textbf{0.386} & \textbf{0.386}& 0.14 &0.028 &0.081 & 0.49\\
        Entity Extraction& \textbf{0.096} & \textbf{0.096}& 0.009  & \textbf{0.096}&0.021  &0.078\\
       Question Answering & \textbf{0.395}& \textbf{0.395} & 0.307&0.31 &0.307  &0.462\\
        English Understanding  & \textbf{0.56} & \textbf{0.56}& 0.493 & 0.404&0.405  &0.56\\
        Semantic Analysis & \textbf{0.549} & \textbf{0.549} & 0.437& 0.337&0.451  &0.549\\
        Syntactic Analysis & 0.569 & \textbf{0.577}& 0.564& 0.547&0.533& 0.577 \\
        Commonsense Morality Understanding  & \textbf{0.805} & \textbf{0.805} &  0.778& 0.792&0.751& 0.808\\\midrule
        Average  & \textbf{0.491}
        & 0.489 & 0.403 &0.394 &0.395 & 0.512\\
        \bottomrule\\
    \end{tabular}}
    \caption{Performance of Different Dataset Selection Methods on Language Tasks}
    \label{tab:language_accuracy}
\end{table}

\textbf{Merge to Mix yielded the best performance}. The first two columns in Table \ref{tab:vision_accuracy} and \ref{tab:language_accuracy} correspond to the performance of our selection method Merge to Mix. In column 1, we report the test accuracy of the selected merged model  $\operatorname{Acc}(\tilde\btheta(\mathcal{S}_{\balpha^*}); T)$, whereas in column 2, we report the accuracy of the corresponding mixture-fine-tuned model $\operatorname{Acc}(\btheta^*(\mathcal{S}_{\balpha^*});T)$. The next three columns correspond to  baselines, reporting the test accuracy of models fine-tuned on (\rm i) all available training datasets combined, (\rm ii) mixtures of datasets selected using a similarity-based metric, and (\rm iii) randomly selected mixtures of datasets.

In both tables, Merge to Mix nearly always achieves the highest accuracy, often approaching or even matching oracle performance. Note that in some cases Merge to Mix exceeds oracle performance on the test set. This is because, as for all columns in the table, the oracle's best mixture was selected based on validation set performance. While using all datasets or relying on similarity metrics are common heuristics, our results show that neither were able to match the performance of Merge to Mix. Interestingly, the best similarity-based approach performed no better than simply finetuning on all datasets. This is particularly surprising given that, to favor the comparison for similarity selection, Tables~\ref{tab:vision_accuracy} and~\ref{tab:language_accuracy} report only the metric that achieved the highest average accuracy among six similarity-based metrics described in Appendix~\ref{appendix:similarity_based_selection}. For both image classification tasks (Table~\ref{tab:vision_accuracy}) and language tasks (Table~\ref{tab:language_accuracy}), the best-performing similarity-based metric was the minimum of minimum $L_2$ distance (Equation \ref{min_of_min_l2}), with an average performance of 0.483 and 0.394, respectively. In Appendix \ref{appendix:additional_eval_results}, we provide a comprehensive comparison of all six metrics across all tasks. Random selection, which serves as a non-heuristic baseline that requires no prior knowledge, also performed poorly. These results show that approaches like maximizing the amount of training data or using embedding-based dataset similarity do not perform well at selecting the optimal mixture of datasets.

\textbf{Merged models sometimes outperformed mixture-fine-tuned models.} Table \ref{tab:vision_accuracy} suggests that, in the image classification tasks, merged models often outperformed even the oracle, achieving an average accuracy that was $6.3\%$ higher than that of mixture-fine-tuned models. This trend was not evident for the language tasks from Table \ref{tab:language_accuracy}. Notably, for image classification tasks Cars, DTD, RESISC45 and SUN397, fine-tuning on any dataset or dataset mixture consistently reduced the model's performance relative to the pre-trained model (see Appendix \ref{appendix:additional_eval_results} for zero-shot accuracy of the pre-trained model). In contrast, merged models were far less susceptible to this degradation, with often all merged models outperforming their mixture-fine-tuned counterparts and, in some cases, even the original pre-trained model. These phenomena raise the intriguing question of why merged models better resist fine-tuning that harms the target task. While model merging has previously been used to improve robustness \cite{wortsman2022robust, wortsman2022model, rame2022diverse, izmailov2018averaging}, it remains unclear why this enhanced resilience exists.

\section{Conclusion and Future Directions}\label{conclusion}
We introduced Merge to Mix, a novel and effective method for dataset mixture selection. By leveraging the correlation between mixture-fine-tuned and merged model performance, our approach identified high-performing dataset mixtures, without requiring fine-tuning for every candidate mixture. Across multiple tasks and modalities, Merge to Mix outperformed the accuracy of standard approaches. This work opens several research directions that we leave for future exploration. Below, we highlight three of the most promising avenues.

First, our method can be combined with search algorithms or predictive models to improve scalability as the number of available training datasets increases. In our experiments (Section~\ref{experiments}), the number of candidate datasets per target task is relatively small ($N \in \{6, 7\}$). In this case, the use of the surrogate made it feasible to exhaustively evaluate all $2^N$ possible dataset mixtures. However, as $N$ grows, this brute-force approach becomes computationally prohibitive despite the fact that Merge to Mix eliminates the training cost associated evaluating a candidate mixture. In such scenarios, efficient search algorithms or predictive models can be readily integrated with our approach to search or predict promising candidate mixtures, enabling effective exploration of the exponential solution space without requiring exhaustive evaluation.

Second, as discussed in Section \ref{sec:review_on_data_selection}, our method can serve as a complementary method for selection approaches that rely on proxy or predictive models. For proxy models~\cite{xie2023doremi, chen2023alpagasus, fan2023doge, coleman2019selection},  Merge to Mix can be used to further enhance scalability by avoiding the need to fine-tune the proxy model on each candidate mixture. For predictive models~\cite{aharoni2020unsupervised, iter2021complementarity, gunasekar2023textbooks, engstrom2024dsdm, liu2024regmix}, Merge to Mix can be used to generate a larger number of  mixture performance evaluations for training a more accurate predictive model. 

Third, as mentioned in Section \ref{sec:algorithm}, Merge to Mix can be extended to weighted dataset selection where $\balpha\in [0,1]^N$. Our framework can naturally accommodate this extension through several strategies. For example, by including multiple copies of the datasets in the candidate pool, Merge to Mix can select datasets with different frequencies, effectively simulating different weights. Alternatively, merging techniques such as weighted averaging can be employed to reflect the relative importance of datasets by adjusting the weights of the corresponding models during merging. Additionally, alternative merging methods, such as~\cite{ilharco2022editing, yadav2023ties}, are worth investigating as standalone techniques for their potential impact on selection outcomes.

\subsubsection*{Author Contributions}
K. Vinken ideated the research; K. Vinken and X. Boix conceptualized the algorithm; Z. Tao conceptualized the theoretical part with contributions from K. Vinken and X. Boix; K. Vinken conceptualized the experimental part with contributions of Z. Tao and X. Boix; Z. Tao, K. Vinken, A. Cooper and H. Yeh wrote the code and ran the experiments; K. Vinken analyzed the experimental results with contributions from Z. Tao and X. Boix; Z. Tao wrote the paper with contributions from K. Vinken and X. Boix; X. Boix supervised the project. 
\bibliographystyle{plain}
\bibliography{myref.bib}


\appendix

\newpage
\section{Additional Experimental Setup Details}\label{appendix:additional_experiment_setup}
In this section, we provide additional details on our experimental setup to complement Section \ref{experimental_setup}.
\subsection{Compute Resources}
All experiments on image classification tasks were carried out on a server equipped with 8 NVIDIA A100 GPUs. For language tasks, all training and evaluation was carried out on 8 NVIDIA A100 80GB GPUs, whereas training and test sample embeddings were computed using 4 NVIDIA A100 40GB GPUs.
\subsection{Datasets and Evaluation Metrics for Language Tasks}\label{appendix:ds_and_eval_for_language_tasks}
We report additional details on the datasets and evaluation metrics used for language tasks. Table~\ref{tab:training_datasets_language} lists the training datasets, while Table~\ref{tab:test_datasets_metrics_language} presents the test datasets along with their corresponding evaluation metrics. We used a battery of benchmarks with different metrics to evaluate Japanese skills in addition to English understanding. Please refer to \url{https://github.com/llm-jp/llm-jp-eval} and \url{https://github.com/wandb/llm-leaderboard} for additional information on benchmarks. 
\begin{table}[H]
    \centering
    \resizebox{1.0\textwidth}{!}{%
    \begin{tabular}{ccc}
        \toprule
 \textbf{Dataset}&\textbf{Description}&\textbf{Number of Samples $N$}\\\midrule
 mCoT-MATH-JA \tablefootnote{\url {https://huggingface.co/datasets/laihuiyuan/mCoT-MATH}} & \makecell{machine-translated Japanese \\math CoT reasoning dataset}&\makecell{$N = 50,000$\\ first 50,000 of the Japanese samples\\(column = `\texttt{lang}', value = `\texttt{ja}')}\\\midrule\\
 oasst2-33k-ja \tablefootnote{\url{https://huggingface.co/datasets/llm-jp/oasst2-33k-ja}} & \makecell{a machine-translated Japanese of \\an English subset from oasst2 \tablefootnote{\url {https://huggingface.co/datasets/OpenAssistant/oasst2}}}&\makecell{$N = 32,701$\\ all samples}\\\midrule\\
 oasst1-21k-en \tablefootnote{\url{https://huggingface.co/datasets/llm-jp/oasst1-21k-en}}&  an English subset of oasst1 \tablefootnote{\url{https://huggingface.co/datasets/OpenAssistant/oasst1}}&\makecell{$N = 21,162$\\all samples}\\\midrule\\
 Aya-Dataset-Japanese \tablefootnote{\url {https://huggingface.co/datasets/CohereForAI/aya_dataset}}& \makecell{Japanese human-annotated \\prompt completion pairs}&\makecell{$N = 6,259$\\only the Japanese samples \\ (column = `\texttt{language\_code}', value = `\texttt{jpn}')}\\\midrule\\
 orca-translations& \makecell{Japanese-English machine-translated\\ samples from ultra-orca-boros-en-ja-v1 \tablefootnote{\url{https://huggingface.co/datasets/augmxnt/ultra-orca-boros-en-ja-v1}}}&\makecell{$N = 14, 103$\\all samples labeled as translations \\(column = `\texttt{source}', value = `\texttt{translations}')}\\\midrule\\
 orca-slimorca & \makecell{slimorca samples \\from ultra-orca-boros-en-ja-v1 \tablefootnote{\url{https://huggingface.co/datasets/augmxnt/ultra-orca-boros-en-ja-v1}}}&\makecell{$N = 12,020$\\all samples labeled as slimorca\\(column = `\texttt{source}', value = `\texttt{slimorca}')}\\
        \bottomrule\\
    \end{tabular}}
    \caption{Training Datasets for Language Tasks}
    \label{tab:training_datasets_language}
\end{table}




\begin{table}[ht]
    \centering
    \resizebox{1.0\textwidth}{!}{%
    \begin{tabular}{ccc}
        \toprule
 \textbf{Task}&\textbf{Benchmark}&\textbf{Metric}\\\midrule
 \multirow{4}{*}{Translation}& alt-e-to-j&\multirow{2}{*}{bleu-ja}\\
 &wikicorpus-e-to-j&\\
 &alt-j-to-e&\multirow{2}{*}{bleu-en}\\
 &wikicorpus-j-to-e&\\\midrule
 Information Extraction &isquad&char-f1\\\midrule
 Mathematical Reasoning & mawps&exact-match-figure \\\midrule
 Entity Extraction &chabsa&set-f1\\\midrule
 \multirow{5}{*}{Question Answering}&jcommonsenseqa&\multirow{2}{*}{exact-match}\\
 &jmmlu&\\
 &jemhopga&\multirow{3}{*}{char-f1}\\
 &niilc&\\
 &aio&\\\midrule
 English Understanding & mmlu-en&exact-match\\\midrule
 \multirow{5}{*}{Semantic Analysis}&jnli&\multirow{5}{*}{exact-match}\\
 &janli&\\
 &jsem&\\
 &jsick&\\
 &jamp&\\\midrule
 \multirow{4}{*}{Syntactic Analysis}&wiki-reading&char-f1\\
 &jblimp&\multirow{3}{*}{exact-match}\\
 &jcola-in-domain&\\
 &jcola-out-of-domain&\\\midrule
 Commensense Morality Understanding &commonsensemoralja&exact-match\\
        \bottomrule\\
    \end{tabular}}
    \caption{Test Datasets and Metrics for Language Tasks}
    \label{tab:test_datasets_metrics_language}
\end{table}
\subsection{Fine-Tuning}\label{appendix:details_on_fine_tuning}
For image classification tasks, we fine-tuned CLIP-ViT-B-32 on each training dataset for 10 epochs using a batch size of 128 and a learning rate of $1\text{e}{-5}$. For language tasks, we fine-tuned Llama-3-8B-Instruct for 1 epoch with a learning rate of 0.01 and a context length of 6000 tokens. Fine-tuning in language tasks was performed using the IA3 method \cite{liu2022few}.

\subsection{Similarity-Based Selection}\label{appendix:similarity_based_selection}
In this section, we present six similarity-based metrics used as baselines. To compute a similarity score between a target task $T$ and a dataset mixture $\mathcal{S}$, we consider six variants of cosine similarity and $L_2$ distance:
(1) average of maximum cosine similarity, (2) average of minimum $L_2$ distance, (3) average of average cosine similarity, (4) average of average $L_2$ distance, (5) maximum of maximum cosine similarity, and (6) minimum of minimum $L_2$ distance. These are formally defined as follows.

\begin{align}\label{avg_of_max_cos}
    \cos(T;\mathcal{S}) =\frac{1}{|T|}\sum_{x\in T}\max_{y\in \mathcal{S}}\frac{x\cdot y}{\|x\|\|y\|},
\end{align}
\begin{align}\label{avg_of_min_L2}
    L_2(T;\mathcal{S}) = \frac{1}{|T|}\sum_{x\in T}\min_{y\in \mathcal{S}}\|x-y\|.
\end{align}
\begin{align}\label{avg_of_avg_cos}
    \cos(T;\mathcal{S}) =\frac{1}{|T|}\sum_{x\in T}\frac{1}{|\mathcal{S}|}\sum_{y\in \mathcal{S}}\frac{x\cdot y}{\|x\|\|y\|}
\end{align}
\begin{align}\label{avg_of_avg_l2}
    L_2(T;\mathcal{S}) = \frac{1}{|T|}\sum_{x\in T}\frac{1}{|\mathcal{S}|}\sum_{y\in \mathcal{S}}\|x-y\|
\end{align}
\begin{align}\label{max_of_max_cos}
    \cos(T;\mathcal{S}) =\max_{x\in T}\max_{y\in \mathcal{S}}\frac{x\cdot y}{\|x\|\|y\|}
\end{align}
\begin{align}\label{min_of_min_l2}
    L_2(T;\mathcal{S}) = \min_{x\in T}\min_{y\in \mathcal{S}}\|x-y\|
\end{align}
When computing similarity-based metrics, we first extracted embeddings using the same pre-trained models that were also used in the fine-tuning experiments: CLIP-ViT-B-32 for image classification tasks and the final transformer layer of Llama-3-8B-Instruct for language tasks. We then applied the formula described above to these embeddings to obtain the final scores.

For image classifications, we used embeddings from the validation splits of both the target tasks and the training datasets. For language tasks, we computed embeddings from  the validation splits of target tasks and full training datasets. To generate these embeddings, we used the full sample, including the assistant tokens. To get a single embedding vector per sample, we first masked out special tokens and tokens outside the attention mask, and then averaged across remaining tokens.

\section{Additional Experiment Results and Plots}\label{appendix:additional_exp_results_and_plots}
This section provides additional experimental results and plots that complement Section \ref{experiments}.

\subsection{Additional Correlation Plots}\label{appendix:correlation_plots}
\begin{figure}[htbp]
    \centering
    \includegraphics[width=0.9\linewidth]{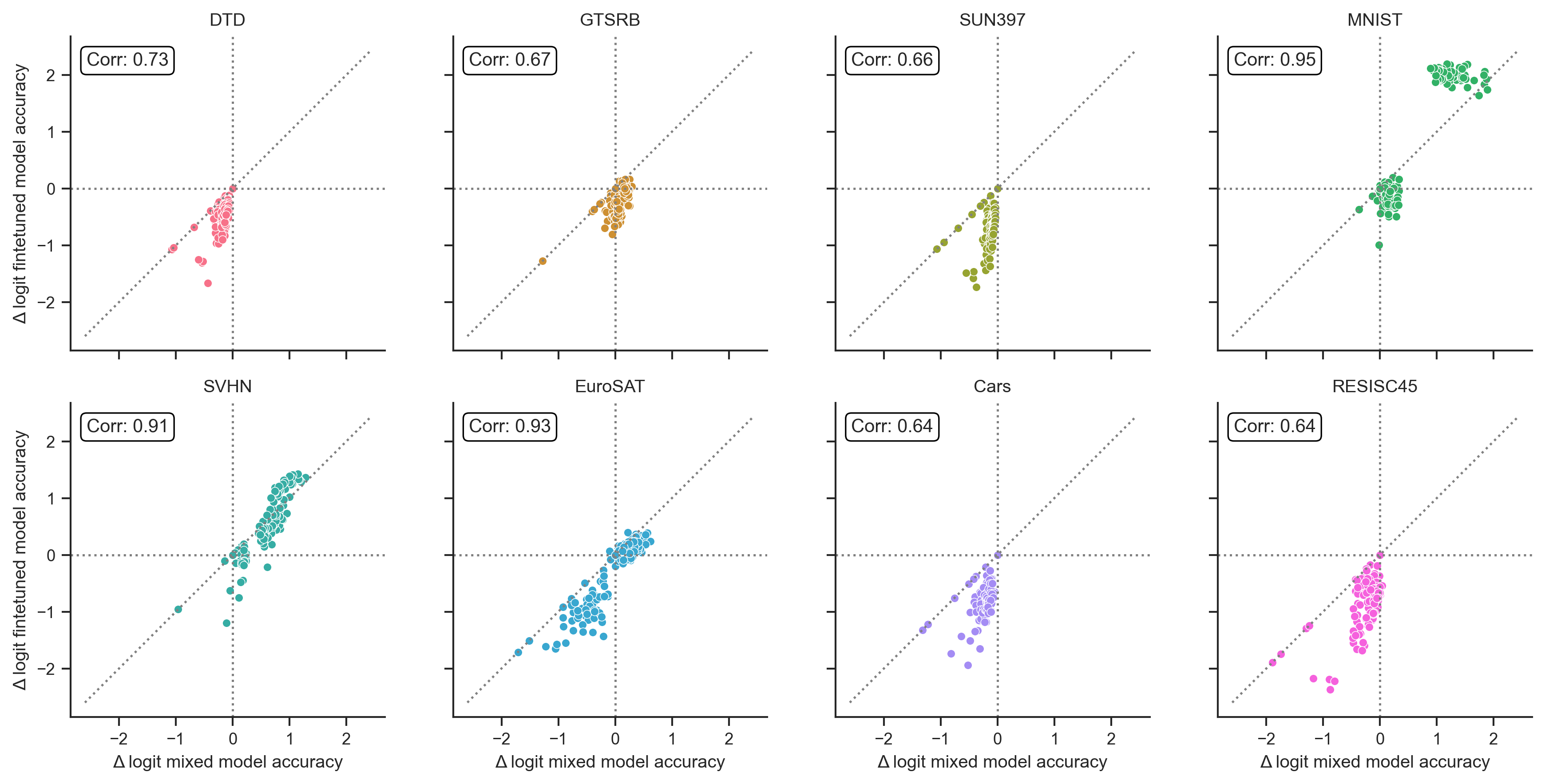}
    \caption{Individual correlation plot between merged model and mixture-fine-tuned model for each target image classification task.}
\label{fig:individual_correlation_plot_merged_model_vision}
\end{figure}

\begin{figure}[htbp]
    \centering
    \includegraphics[width=0.9\linewidth]{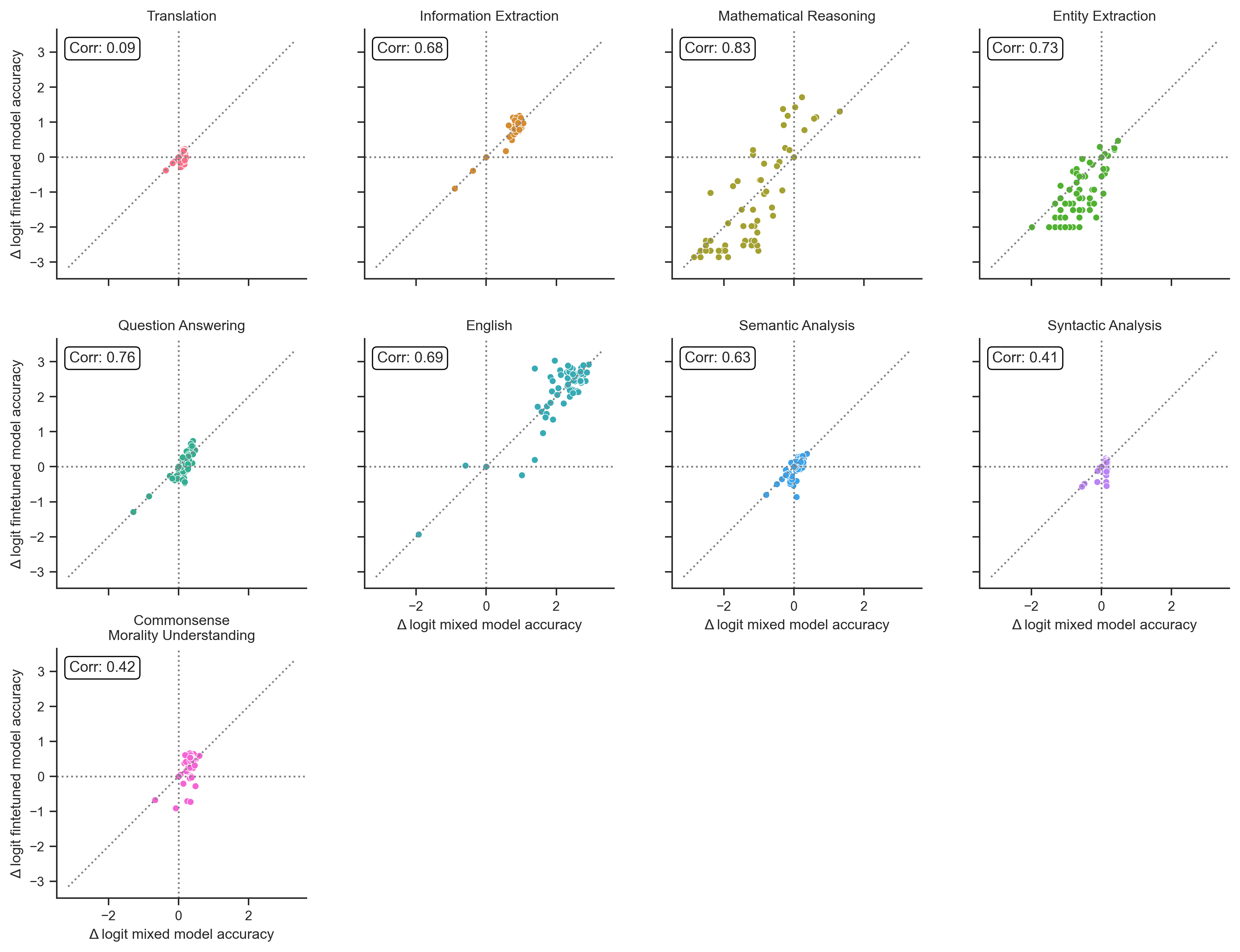}
    \caption{Individual correlation plot between the performance of merged models and mixture-fine-tuned models for each target language task.}
\label{fig:individual_corrrelation_plot_merged_model_language}
\end{figure}

\begin{figure}[htbp]
\begin{subfigure}{.5\textwidth}
    \centering
    \includegraphics[width=0.9\linewidth]{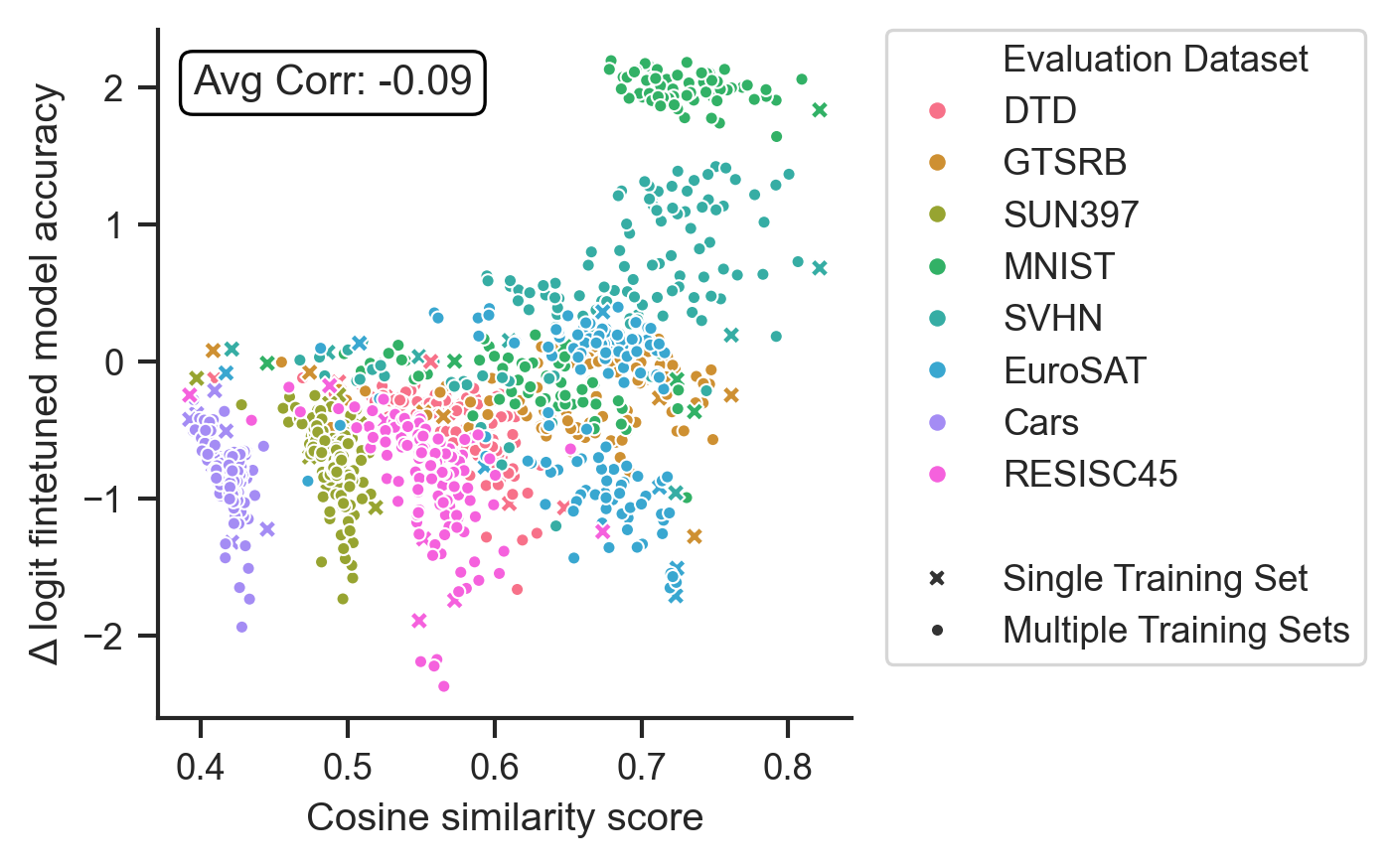}
    \caption{Image Classification}
\label{fig:correlation_avg_of_avg_cosine_vision}
\end{subfigure}
  \begin{subfigure}{.5\textwidth}
    \centering
    \includegraphics[width=0.9\linewidth]{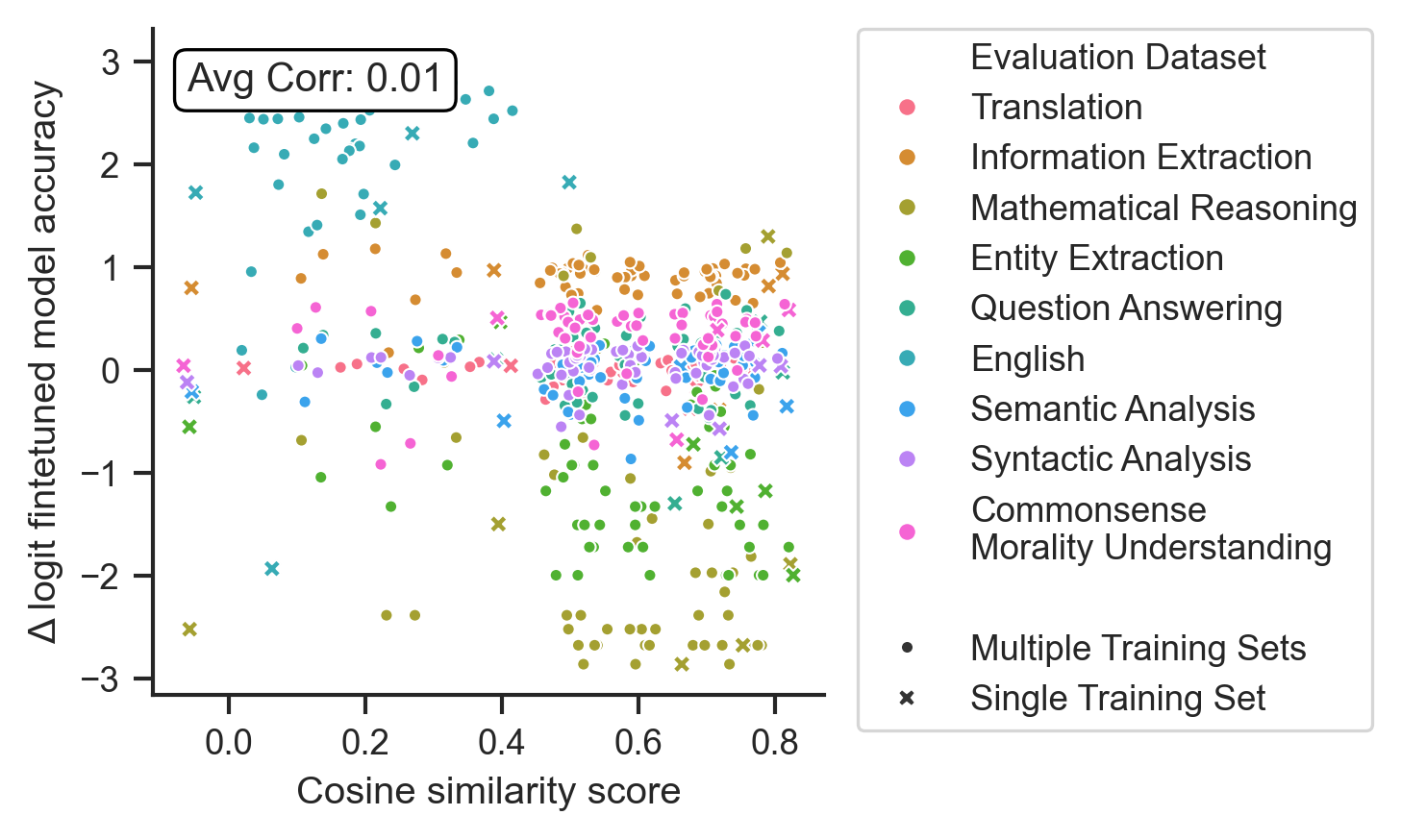}
    \caption{Language}
\label{fig:correlation_avg_of_avg_cosine_language}
\end{subfigure}
\caption{Correlation plots between the performance of the mixture-fine-tuned models and average of average cosine similarity metric.}
\label{fig:correlation_plot_avg_of_avg_cosine}
\end{figure}

\begin{figure}[htbp]
\begin{subfigure}{.5\textwidth}
    \centering
    \includegraphics[width=0.9\linewidth]{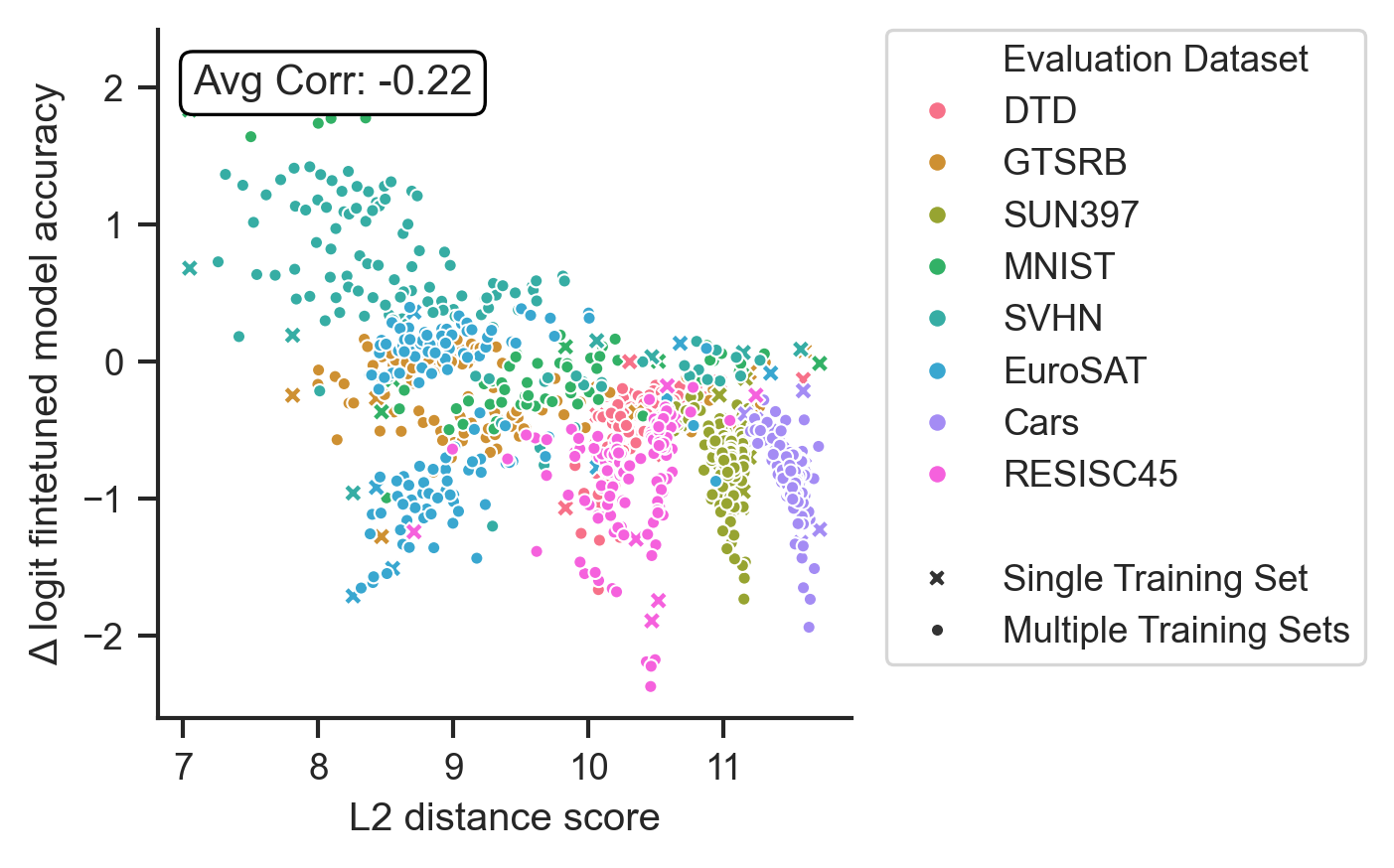}
    \caption{Image Classification}
\label{fig:correlation_avg_of_avg_L2_vision}
\end{subfigure}
  \begin{subfigure}{.5\textwidth}
    \centering
    \includegraphics[width=0.9\linewidth]{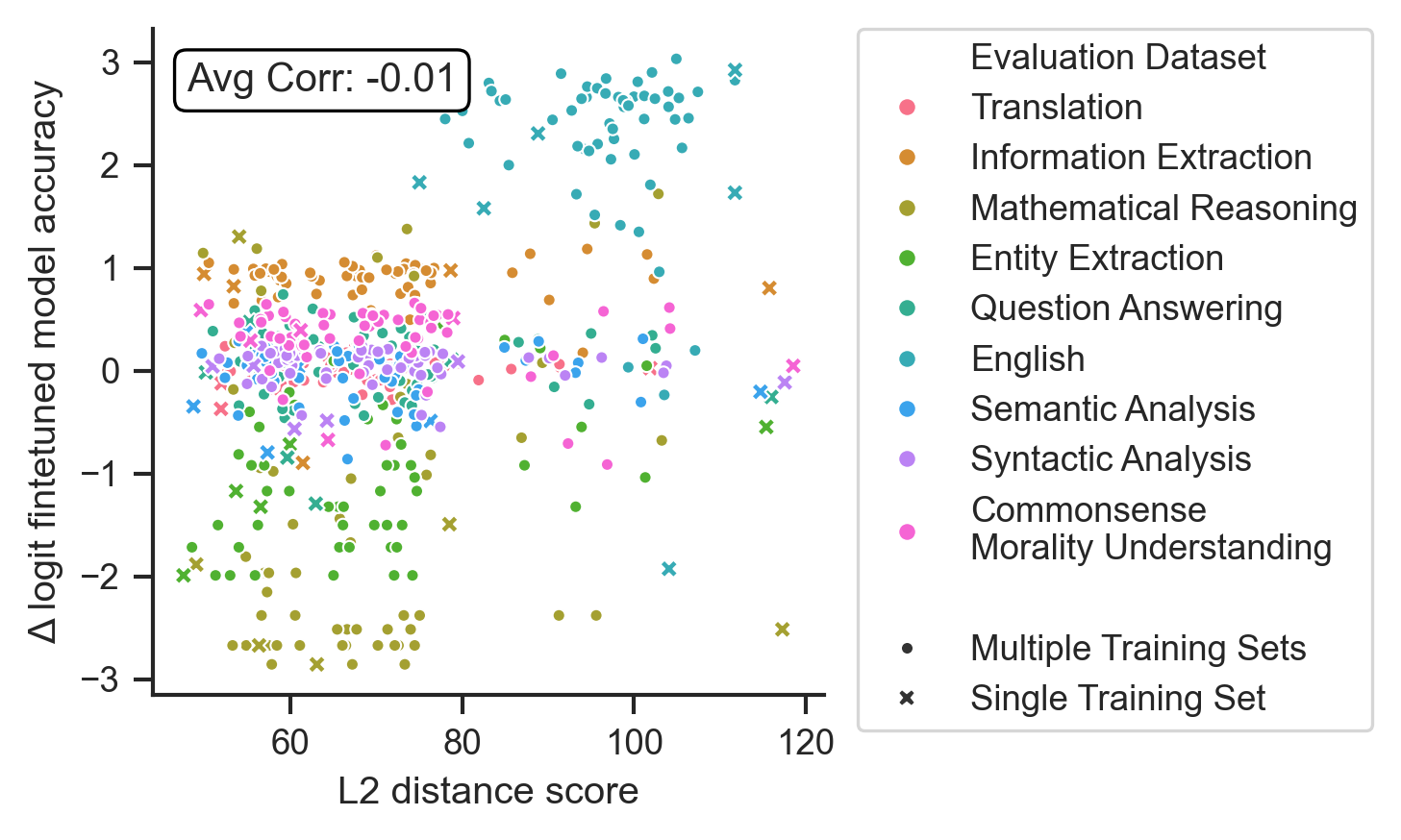}
    \caption{Language}
\label{fig:correlation_avg_of_avg_L2_language}
\end{subfigure}
\caption{Correlation plots between the performance of the mixture-fine-tuned models and average of average $L_2$ score.}
\label{fig:correlation_plot_avg_of_avg_L2}
\end{figure}

\begin{figure}[htbp]
\begin{subfigure}{.5\textwidth}
    \centering
    \includegraphics[width=0.9\linewidth]{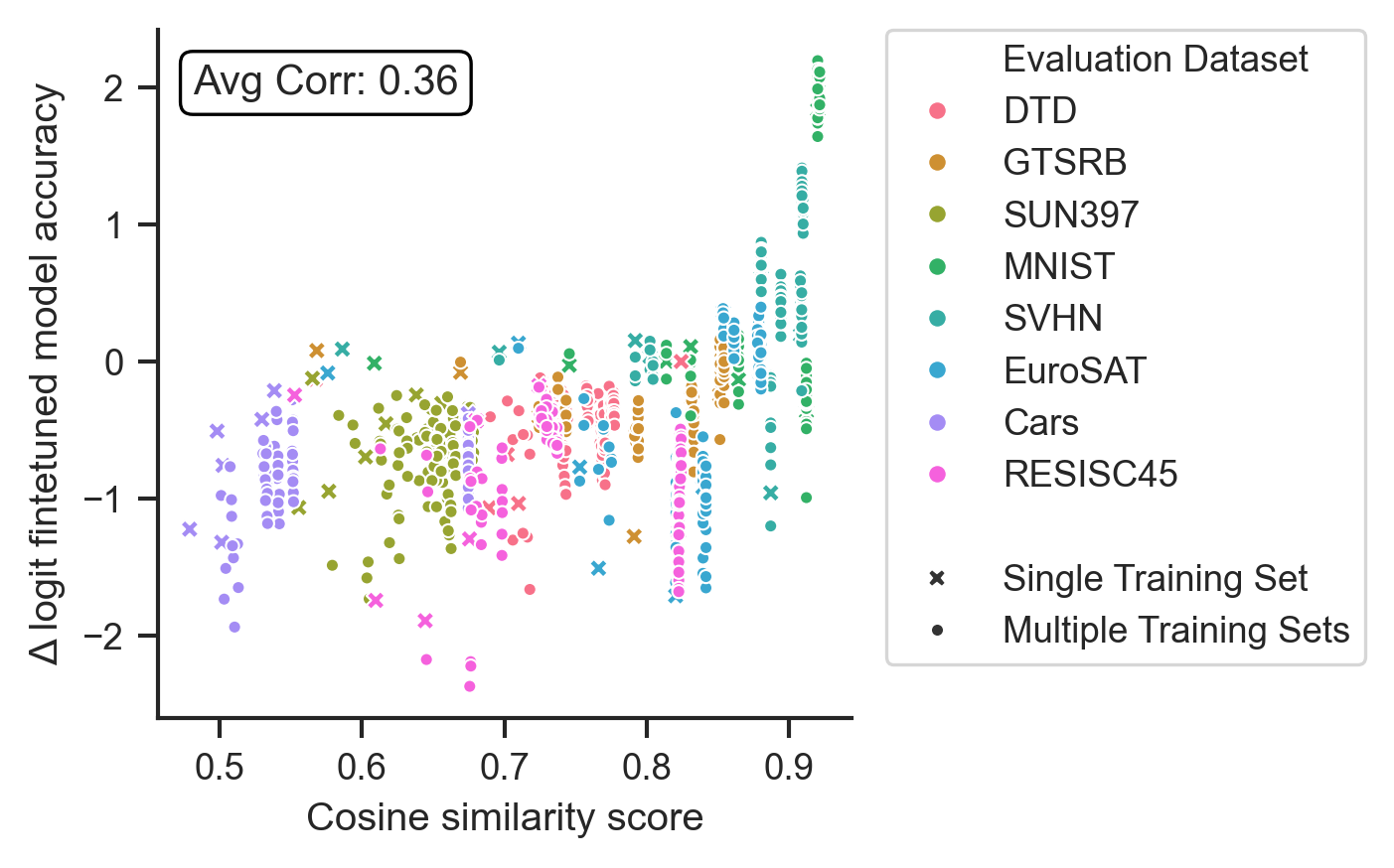}
    \caption{Image Classification}
\label{fig:correlation_max_of_max_cos_vision}
\end{subfigure}
  \begin{subfigure}{.5\textwidth}
    \centering
    \includegraphics[width=0.9\linewidth]{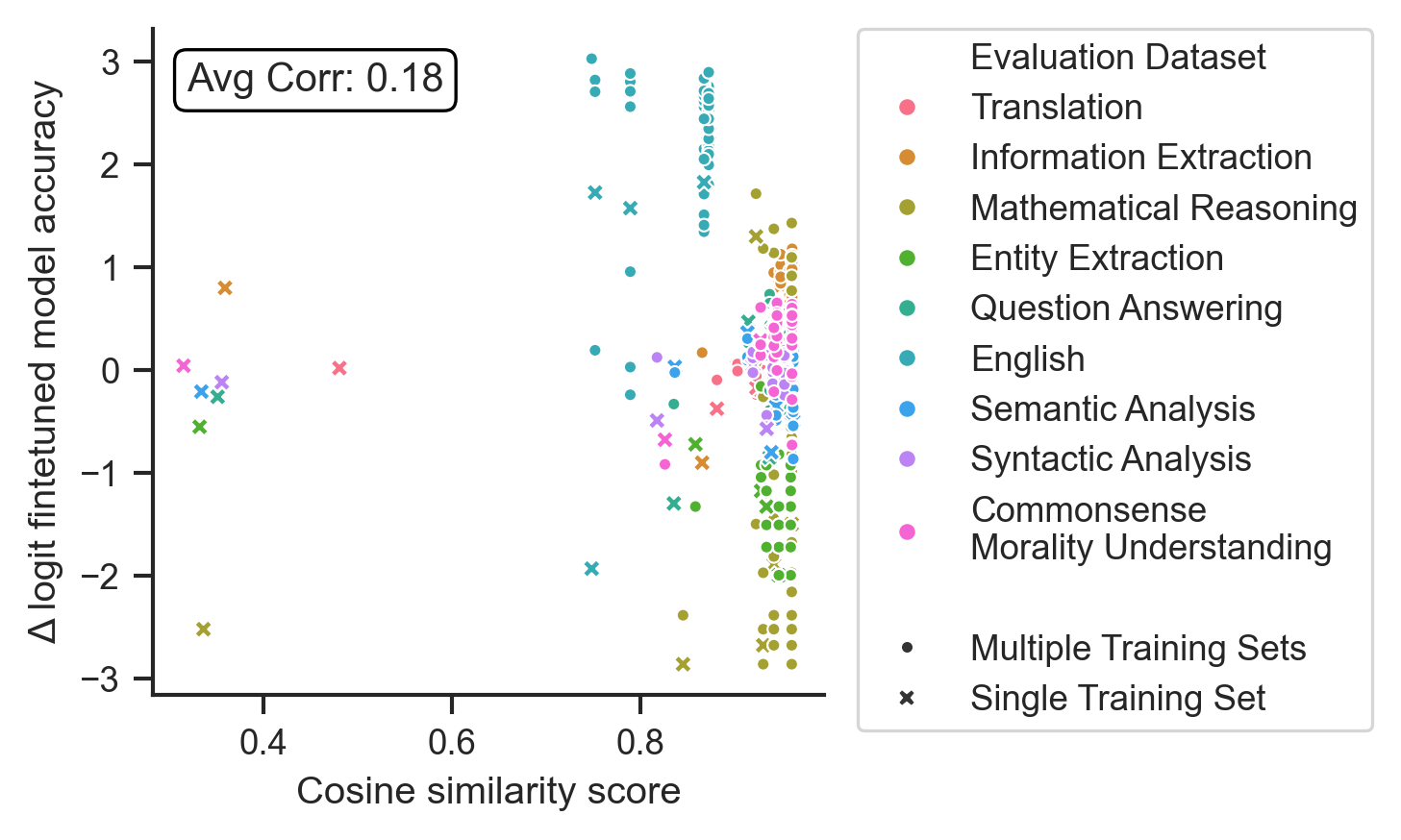}
    \caption{Language}
\label{fig:correlation_max_of_max_cos_language}
\end{subfigure}
\caption{Correlation plots between the performance of the mixture-fine-tuned models and maximum of maximum cosine similarity.}
\label{fig:correlation_plot_max_of_max_cos}
\end{figure}

\begin{figure}[htbp]
\begin{subfigure}{.5\textwidth}
    \centering
    \includegraphics[width=0.9\linewidth]{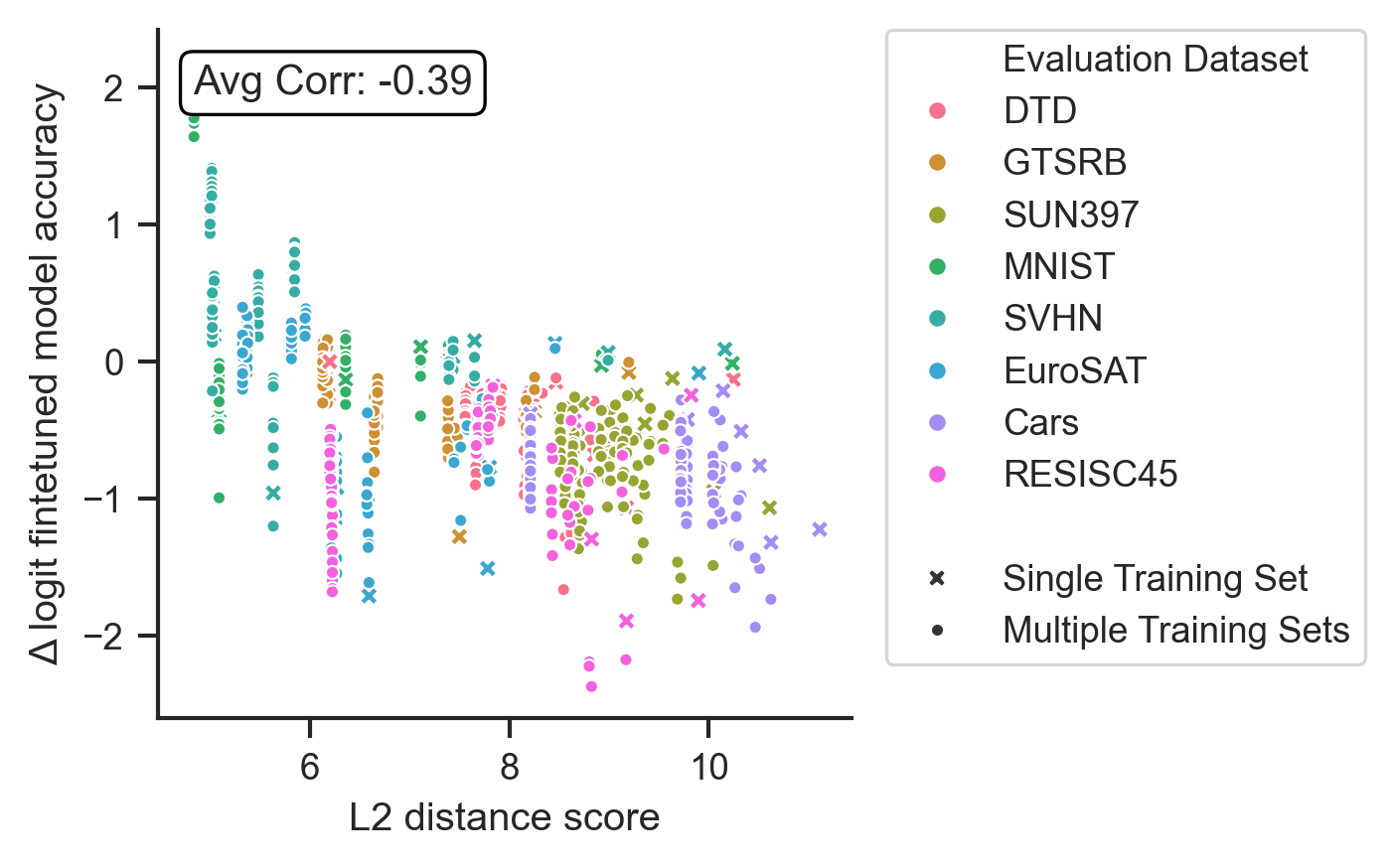}
    \caption{Image Classification}
\label{fig:correlation_min_of_min_L2_vision}
\end{subfigure}
  \begin{subfigure}{.5\textwidth}
    \centering
    \includegraphics[width=0.9\linewidth]{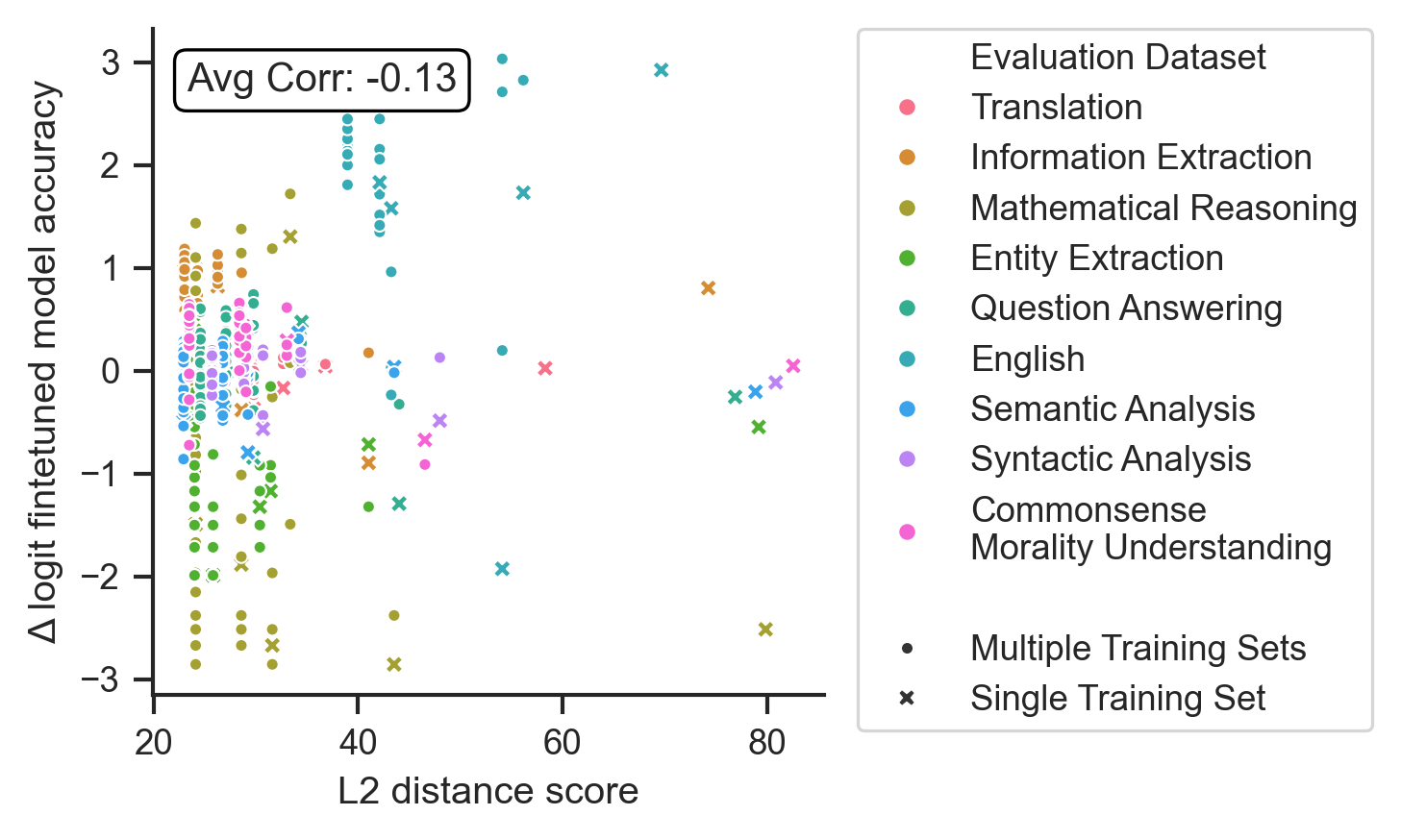}
    \caption{Language}
\label{fig:correlation_min_of_min_L2_language}
\end{subfigure}
\caption{Correlation plots between the performance of the mixture-fine-tuned models and minimum of minimum $L_2$ score.}
\label{fig:correlation_plot_min_of_min_L2}
\end{figure}

We present correlation plots for each individual target task along with the corresponding Pearson correlation coefficients. Figure~\ref{fig:individual_correlation_plot_merged_model_vision} shows the correlation between the performance of the merged model and the mixture-fine-tuned model for each image classification task, while Figure~\ref{fig:individual_corrrelation_plot_merged_model_language} presents the same for each language task. Figures~\ref{fig:correlation_plot_avg_of_avg_cosine}, \ref{fig:correlation_plot_avg_of_avg_L2}, \ref{fig:correlation_plot_max_of_max_cos}, and~\ref{fig:correlation_plot_min_of_min_L2} provide additional correlation plots between the performance of the mixture-fine-tuned models and the similarity-based metrics computed using Equations~\eqref{avg_of_avg_cos}, \eqref{avg_of_avg_l2}, \eqref{max_of_max_cos}, and~\eqref{min_of_min_l2}, respectively.


\subsection{Additional Evaluation Results}\label{appendix:additional_eval_results}
We present additional evaluation results for comprehensive understanding of our results in Section \ref{experiments}. Table \ref{tab:all_six_similarity_accuracy_vision} and \ref{tab:all_six_similarity_accuracy_language} report the performance of all six similarity-based metrics described in Section \ref{appendix:similarity_based_selection} for image classification and language tasks, respectively. We also report the zero-shot accuracy of the pre-trained models used for each target task in Table~\ref{tab:zeroshot_acc_of_vision} and Table~\ref{tab:zeroshot_acc_of_language}. Specifically, we use ViT-B-32 for image classification tasks and Llama-3-8B-Instruct for language tasks.
\begin{table}[htbp]
    \centering
    \resizebox{1.0\textwidth}{!}{%
    \begin{tabular}{ccccccc}
        \toprule
        \diagbox{\textbf{Target Task}}{\textbf{Selection Method}}
 & \makecell{\textbf{Average of Average}\\\textbf{Cosine Similarity}} &\makecell{\textbf{Average of Average}\\\textbf{$L_2$ Distance}} &\makecell{\textbf{Max of Max}\\\textbf{Cosine Similarity}}& \makecell{\textbf{Min of Min}\\\textbf{$L_2$ Distance}}&\makecell{\textbf{Average of Max}\\\textbf{Cosine Similarity}}&\makecell{\textbf{Average of Min}\\\textbf{$L_2$ Distance}} \\
        \midrule
        Cars  & 0.303 & 0.502 &0.47 & 0.456 & 0.502 &0.502\\
        DTD   & 0.215 &0.215 & 0.332 & 0.332 & 0.381 &0.381\\
        EuroSAT  & 0.154 &0.13 &0.429 & 0.429 &0.543 & 0.543\\
        GTSRB & 0.275 &0.275 &0.263 &0.309 & 0.275 & 0.275\\
       MNIST  & 0.854&0.854 &0.863 &0.863 &0.854 &0.854\\
        RESISC45  & 0.305 &0.305 &0.437 &0.445&0.305&0.305\\
        SUN397  & 0.371 & 0.558 & 0.434 &0.469&0.558&0.558\\
        SVHN & 0.478 & 0.478 &0.561 & 0.561&0.359&0.359\\\midrule
        Average &  0.369& 0.415 & 0.474 &0.483&0.472&0.472\\
        \bottomrule\\
    \end{tabular}}
    \caption{Performance of Six Similarity Selection Methods on Image Classification Tasks}
    \label{tab:all_six_similarity_accuracy_vision}
\end{table}
\begin{table}[htbp]
    \centering
    \resizebox{1.0\textwidth}{!}{%
    \begin{tabular}{ccccccc}
        \toprule
 \diagbox{\textbf{Target Task}}{\textbf{Selection Method}}
 & \makecell{\textbf{Average of Average}\\\textbf{Cosine Similarity}} &\makecell{\textbf{Average of Average}\\\textbf{$L_2$ Distance}} &\makecell{\textbf{Max of Max}\\\textbf{Cosine Similarity}}& \makecell{\textbf{Min of Min}\\\textbf{$L_2$ Distance}}&\makecell{\textbf{Average of Max}\\\textbf{Cosine Similarity}}&\makecell{\textbf{Average of Min}\\\textbf{$L_2$ Distance}}\\
        \midrule
        Translation  &  0.159& 0.127&0.184&0.184&0.208&0.208\\
        Information Extraction &0.842 &0.842 &0.847 &0.847 &0.849 &0.839\\
        Mathematical Reasoning  & 0.016 &0.016 &0.028 &0.028 &0.014 &0.014\\
        Entity Extraction& 0.0 &0.0 &0.096 &0.096 &0.019 & 0.019\\
       Question Answering & 0.283 & 0.283 & 0.31 & 0.31 & 0.306 & 0.306\\
        English Understanding & 0.294 & 0.294 & 0.404 & 0.404 & 0.381 & 0.381\\
        Semantic Analysis & 0.37 & 0.37 & 0.337 & 0.337 & 0.337 & 0.337\\
        Syntactic Analysis & 0.535 & 0.535 & 0.547 & 0.547 & 0.571 & 0.571 \\
        Commonsense Morality Understanding & 0.805 & 0.805 & 0.792 & 0.792 & 0.792 & 0.792\\\midrule
        Average  & 0.367 & 0.364 & 0.394 & 0.394 & 0.386 & 0.385\\
        \bottomrule\\
    \end{tabular}}
    \caption{Performance of Six Similarity Selection Methods on Language Tasks}
    \label{tab:all_six_similarity_accuracy_language}
\end{table}

\begin{table}[htbp]
    \centering
    \resizebox{1.0\textwidth}{!}{%
    \begin{tabular}{ccccccccc}
        \toprule
    \textbf{Target Task}&\textbf{Cars}&\textbf{DTD}&
\textbf{EuroSAT}&\textbf{GTSRB}&\textbf{MNIST}&\textbf{RESISC45}&\textbf{SUN397}&\textbf{SVHN}\\\midrule
    \textbf{Zero-shot Accuracy}&0.596&0.444&0.453&0.326&0.483&0.603&0.632&0.316\\
\bottomrule\\
    \end{tabular}}
    \caption{Zero-shot Accuracy of CLIP-ViT-B-32 on Image Classification Tasks}
    \label{tab:zeroshot_acc_of_vision}
\end{table}

\begin{table}
    \centering
    \resizebox{1.0\textwidth}{!}{%
    \begin{tabular}{cccccccccc}
        \toprule
    \makecell{\textbf{Target}\\\textbf{Task}}&\textbf{Translation}&\makecell{\textbf{Information}\\\textbf{Extraction}}&\makecell{\textbf{Mathematical}\\\textbf{Reasoning}}&\makecell{\textbf{Entity}\\\textbf{Extraction}}&\makecell{\textbf{Question}\\\textbf{Answering}}&\makecell{\textbf{English}\\\textbf{Understanding}}&\makecell{\textbf{Semantic}\\\textbf{Analysis}}&\makecell{\textbf{Syntactic}\\\textbf{Analysis}}&\makecell{\textbf{Commensense Morality}\\\textbf{Understanding}}\\\midrule
    \textbf{Zero-shot Accuracy}&0.177&0.668&0.14&0.059&0.287&0.055&0.456&0.525&0.692\\
\bottomrule\\
    \end{tabular}}
    \caption{Zero-shot Accuracy of Llama-3-8B-Instruct on Language Tasks}
    \label{tab:zeroshot_acc_of_language}
\end{table}


\newpage

\end{document}